\documentclass{article}

\PassOptionsToPackage{numbers, compress}{natbib}

\usepackage[preprint]{neurips_2025}
\usepackage{subcaption}

\workshoptitle{Scaling Environments for Agents (SEA)}

\usepackage[utf8]{inputenc} %
\usepackage[T1]{fontenc}    %
\usepackage[colorlinks]{hyperref}       %
\usepackage{url}            %
\usepackage{booktabs}       %
\usepackage{amsfonts}       %
\usepackage{nicefrac}       %
\usepackage{microtype}      %
\usepackage{xcolor}         %
\usepackage{amsmath}

\usepackage{mathtools}
\usepackage{amssymb,amsfonts}
\usepackage{graphicx}
\usepackage{booktabs}
\usepackage{microtype}
\usepackage{nicefrac}
\usepackage{algorithm}
\usepackage{algorithmic}
\usepackage{enumitem}
\usepackage{xspace}
\usepackage{wrapfig}

\definecolor{mypurple}{HTML}{7A4988}
\definecolor{mygreen}{HTML}{2ca02c}
\definecolor{myblue}{HTML}{1f77b4}
\hypersetup{urlcolor=myblue,citecolor=myblue,linkcolor=myblue}

\title{\name: Mutual Information Skill Learning for Structured Response Diversity in LLMs}

\author{%
  Anonymous Authors\\
  \texttt{anonymous@domain.edu}
}

\usepackage{amsthm}
\theoremstyle{remark}
\usepackage{multirow}
\usepackage[capitalize,noabbrev]{cleveref}

\newtheoremstyle{plainupright}%
  {\topsep}%
  {\topsep}%
  {\upshape}%
  {}%
  {\bfseries}%
  {.}%
  { }%
  {}%

\theoremstyle{plainupright}
\newtheorem{lemma}{Lemma}

\newtheorem*{definition}{Definition}

\title{\name: Mutual Information Skill Learning for Structured Response Diversity in LLMs}

\DeclareMathOperator{\Var}{Var}
\renewcommand{\and}{%
  \end{tabular}\hfil\penalty 50\hfil%
  \begin{tabular}[t]{@{}c@{}}\ignorespaces
}
\newcommand{\name}{UpSkill\xspace}

\author{%
    \begin{minipage}{\textwidth}
    \centering
    Devan Shah\thanks{Equal contribution.} \quad Owen Yang$^*$ \\
    \vspace{0.3em}
    Daniel Yang \quad Chongyi Zheng \quad Benjamin Eysenbach \\
    \vspace{0.3em}
    \normalfont Princeton University \\
    \vspace{0.5em}
    \texttt{\{devan.shah, owen.yang, daniel.yang, chongyiz, eysenbach\}@princeton.edu}
    \end{minipage}
}

\begin{document}

\maketitle

\begin{abstract}
Reinforcement Learning with Verifiable Rewards (RLVR) has improved the reasoning abilities of large language models (LLMs) on mathematics and programming tasks, but standard approaches that optimize single-attempt accuracy can inadvertently suppress response diversity across repeated attempts, narrowing exploration and overlooking underrepresented strategies.
We introduce \name, a training-time method that adapts \emph{Mutual Information Skill Learning} (MISL) to LLMs for optimizing \texttt{pass@k} performance.
We propose a novel reward that we implement within Group Relative Policy Optimization (GRPO): a \emph{token-level} mutual information (MI) reward that encourages trajectory specificity to $z$. Experiments on GSM8K with three open-weight models, Llama~3.1--8B, Qwen~2.5--7B, and R1-Distilled–Qwen2.5–Math–1.5B, show that \name improves multi-attempt metrics on the stronger base models, yielding mean gains of $\sim$3\% in \texttt{pass@k} for both Qwen and Llama
without degrading \texttt{pass@1}.\footnote[1]{Code: \href{https://github.com/dshah02/upskill}{https://github.com/dshah02/upskill}.} Additionally, we find both empirical and theoretical evidence that improvements in \texttt{pass@k} are closely tied to the mutual information objective. %
The project website can be found at \url{https://dshah.io/upskill}.
\end{abstract}

\section{Introduction}
\label{sec:introduction}

\begin{figure}[th]
    \centering
    \includegraphics[width=0.8\linewidth]{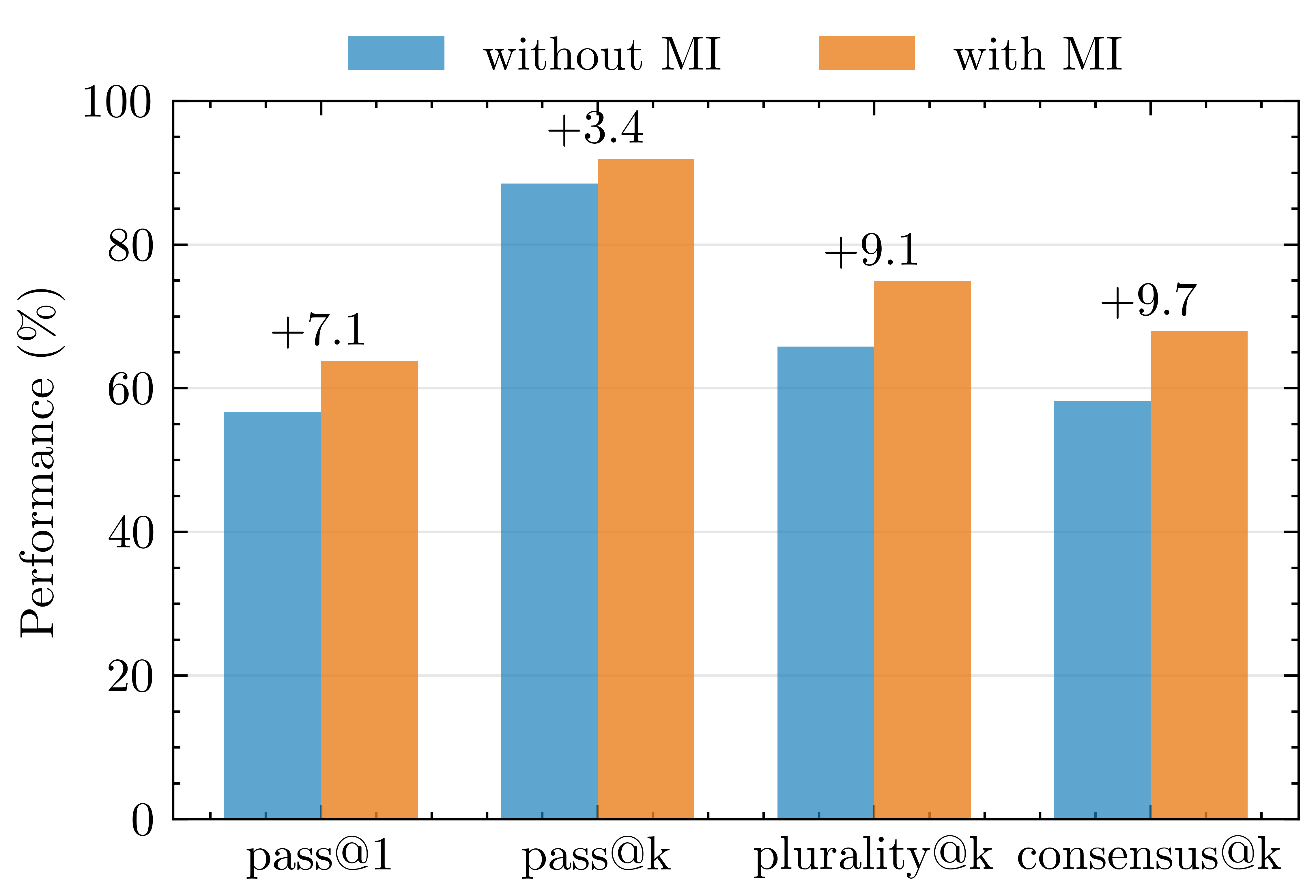}
    \caption{\name improves mean multi-attempt accuracy without hurting single-attempt accuracy on GSM8K for the Qwen 2.5-7B model (See Sec. \ref{sec:gsm}).}
    \vspace{-1.5em}
    \label{fig:qwen_solo_results}
\end{figure}
LLMs excel at verifiable reasoning tasks such as mathematical problem solving and code generation \citep{guo2025deepseek}. However, repeated sampling often yields highly similar outputs \citep{shaier_asking_2025}. This is detrimental in multi-attempt settings where just one correct completion solves the problem at hand, such as code generation with tests \citep{chen_evaluating_2021} or formal proofs in Lean \citep{trinh2024solving}, as a lack of diversity reduces the effective number of independent attempts. Therefore, for these or other objectives evaluated by \texttt{pass@k}, or the probability that at least one of $k$ completions will be correct, greater similarity among outputs reduces the probability of success. Furthermore, recent work has found that post-training that optimizes single-attempt correctness suppresses response variation across attempts \citep{chen2025passktrainingadaptivelybalancing, dang2025assessing}, creating a discrepancy between how models are trained and how they are used and evaluated.

The challenge of balancing diversity and accuracy, or exploration and exploitation~\citep{sutton_reinforcement_2015}, has primarily been studied in prior works that change how decoding is done. Methods such as temperature sampling~\citep{renze_effect_2024}, nucleus sampling~\citep{holtzman2020curiouscaseneuraltext}, and prompt perturbations~\citep{shur-ofry_growing_2024} can inject variety, but they require manual tuning~\citep{du_optimizing_2025} and are brittle across domains~\citep{shi_thorough_2024, qiang_prompt_2024}.
Separately, prior training-time methods do not expose a controllable basis of strategies and require complex training recipes to properly balance exploration and exploitation \citep{tang2025optimizinglanguagemodelsinference, chen2025passktrainingadaptivelybalancing}. We seek a training-time mechanism that \textit{(i)} increases diversity in a controlled manner, \textit{(ii)} produces semantically distinct and reproducible modes of reasoning, and \textit{(iii)} preserves single-attempt verifiable accuracy.

\begin{figure}[h!]
\centering
    \includegraphics[clip, trim=0.5cm 13.5cm 0.3cm 1.3cm, width=0.9\textwidth]{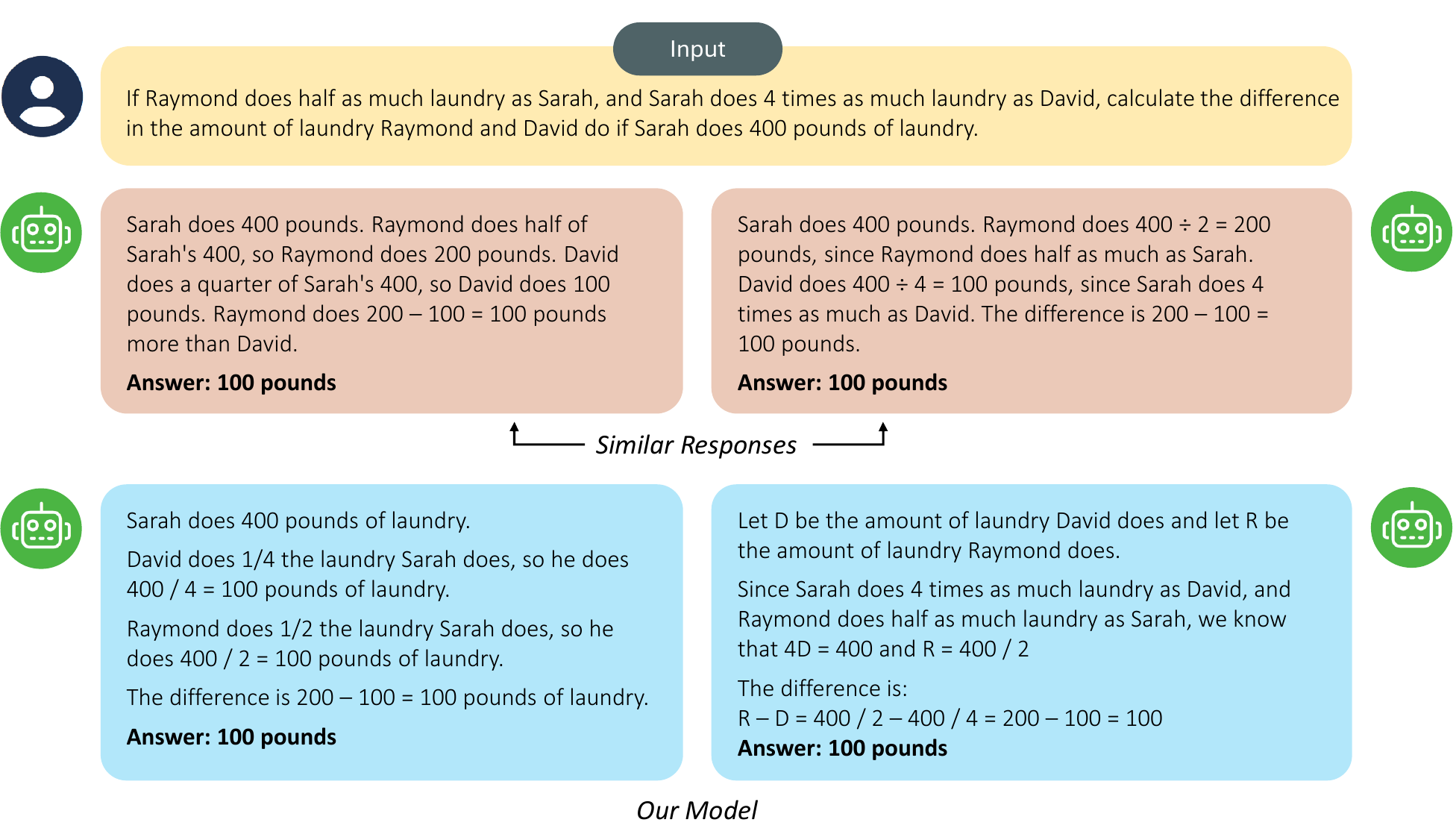}
    \caption{\name is an unsupervised method for training LLMs to produce diverse responses. After training, different latent vectors $z$ (blue boxes above) correspond to different response strategies. Because of space constraints, the figure shows summarized responses from \name; we report the full responses in \autoref{app:generation_figure}.}
    \vspace{-10pt}
    \label{fig:example_generations}
\end{figure}
We introduce \name, a training-time approach that induces \emph{structured response diversity} without prompt engineering.
The key idea behind \name is to introduce an input token $z$ that structures the response, so that different values of $z$ correspond to different responses. Formally, we will model LLM attempts on verifiable reasoning tasks as a token-level Markov decision process. We can then adopt prior work from reinforcement learning on learning \emph{skills}, which learn a policy conditioned on a latent variable $z$. These methods~\citep{eysenbach_diversity_2018, gregor_variational_2016, achiam2018variational, Sharma2020Dynamics-Aware,
florensa2017stochastic} include a loss term that maximizes the mutual information between $z$ and the policy's behavior. Precisely, we adapt the CSF method~\citep{zheng2024can} to LLMs:
the model conditions its response on a discrete latent $z \in \{1, \dots, N\}$, and training encourages behaviors whose distribution depends strongly on $z$. Intuitively, each $z$ should correspond to a reproducible strategy, and the set of strategies should span a broad range of behaviors.

The main contribution of our paper is a method for training LLMs to produce diverse responses. Our method implements mutual information skill learning by applying GRPO \citep{shao2024deepseekmath} with a novel reward term: a token-level mutual information reward, which encourages diversity in completions.
Finally, we sketch a theoretical link between $\mathcal{I}(\tau;z\mid x)$ and \texttt{pass@k}: the improvement in \texttt{pass@k} after training is related to $\mathcal{I}(\tau;z\mid x)$.
 In summary, our contributions are as follows:
\begin{itemize}[itemsep=0pt,topsep=0pt]
    \item \name achieves mean gains of +3.4\% in \texttt{pass@k} and +9.1\% in \texttt{plurality@k} on GSM8K for Qwen 2.5--7B using RL fine-tuning with LoRA adapters on 2,000 problems, with preserved \texttt{pass@1} accuracy. We additionally test two other open-weight models, showing improvement on Llama 3.1--8B and a decline on R1--Distilled--Qwen2.5--Math--1.5B.
    \item We show that, without ground-truth answers, \name can increase \texttt{pass@k} on Qwen 2.5--7B and Llama 3.1--8B.

    \item In an arithmetic puzzle environment, \name improves \texttt{pass@5} by +10\% through mitigating response variation collapse and developing a collection of diverse and complementary skills.
    
    \item We prove that \texttt{pass@k} improvement closely corresponds to the mutual information $\mathcal{I}(\tau;z\mid x)$, showing that large improvements in multi-attempt accuracy require sufficient mutual information.
\end{itemize}
Training and evaluation code is open-sourced at \href{https://github.com/dshah02/upskill}{https://github.com/dshah02/upskill}.

\section{Background and Related Work}
\label{sec:background}

\subsection{Multi-attempt evaluation, redundancy, and why diversity matters}
For verifiable tasks, we often consider the probability of success across \emph{multiple} completions rather than a single attempt \citep{chen2025passktrainingadaptivelybalancing}. Let $x$ denote the input and $\tau$ a sampled completion from policy $\pi(\cdot\mid x)$. Let $Y(\tau)\in\{0,1\}$ indicate correctness under a deterministic verifier. For $k$ attempts, the standard metric
\begin{equation}
\text{pass@}k(x) \;=\; 1 - \Pr\!\left(\bigcap_{i=1}^{k}\{Y(\tau_i)=0\}\,\middle|\,x\right)
\end{equation}
is the complement of the joint failure probability across $k$ i.i.d.\ draws $\tau_{1:k}\sim \pi(\cdot\mid x)$~\citep{chen_evaluating_2021}. 
Letting $p=\Pr(Y(\tau)=1\mid x)$, we therefore have $\text{pass@}k(x)=1-(1-p)^k$.

In practice, identical prompts with fixed decoding hyperparameters can yield strongly correlated trajectories, especially for deterministic or near-deterministic samplers~\citep{vijayakumar_diverse_2018}. A useful lens is to consider an ``effective number of attempts'' $k_{\mathrm{eff}}$ that discounts $k$ by a correlation term (analogous to design effects in sampling)~\citep{kish1965survey}. If completions have pairwise correlation $\rho$ in the binary success indicators, a heuristic adjustment gives $k_{\mathrm{eff}}\approx k/(1+(k-1)\rho)$: as $\rho\!\to\!1$, additional attempts contribute little; as $\rho\!\to\!0$, $k_{\mathrm{eff}}\!\to\!k$. Although crude, this highlights the central point: reducing dependence among attempts is as important as raising per-attempt accuracy. Structured diversity aims to decrease redundancy so that the joint failure probability decreases faster in $k$. For Gaussian random variables, correlation and mutual information are closely related (as intuitively, correlated variables have information on each other) \citep{krafft2013correlation}. However, as text correctness cannot easily be framed as a Gaussian distribution, mutual information is more a natural measure.

Beyond \texttt{pass@k}, \texttt{plurality@k} and \texttt{consensus@k} measure agreement among completions, examining robustness and internal consistency of the model's reasoning~\citep{wallace_estimating_2025}. In many workflows, agreement acts as a proxy for confidence while still benefiting from diversity to escape shared failure modes~\citep{hochlehnert_sober_2025}.

\subsection{RL on language models: token-level MDPs, RLVR, and GRPO}
Autoregressive LLMs can be cast as policies over a Markov decision process (MDP), where the state is the token prefix and the action is the next token~\citep{bahdanau_actor-critic_2017, ouyang_training_2022}. This setup, often referred to as the token-level MDP \citep{zhong2025dpomeetspporeinforced}, allows reinforcement learning algorithms to directly optimize model behavior for correctness on verifiable tasks such as math or code.

Reinforcement Learning from Verifiable Rewards (RLVR) leverages automatically checkable signals (e.g., exact numeric answers, unit tests) as rewards, with the goal being to improve the pass rate of policies while ensuring the new policy remains close to a base model~\citep{liu_rltf_2023, dou_stepcoder_2024}. Let $\pi_\theta$ denote the trainable policy and $\pi_{\mathrm{base}}$ a frozen reference. A common form of the per-trajectory reward is
\begin{equation}
r_{\mathrm{RLVR}}(\tau) \;=\; r_{\text{correctness}}(\tau)\;-\;\beta\, D_{\mathrm{KL}}\!\left(\pi_\theta(\cdot\mid x)\,\big\|\,\pi_{\mathrm{base}}(\cdot\mid x)\right),
\end{equation}
where $\beta>0$ controls deviation from the base model~\citep{xiong_iterative_2024}.

Group Relative Policy Optimization (GRPO) \citep{shao2024deepseekmath} adapts PPO-style updates to reasoning by sampling multiple completions per prompt $x$ as a \emph{group}. Within-group baselines reduce variance and increase the relative difference between completion rewards. Concretely, for each $x$ one draws $C$ trajectories $\{\tau_i\}_{i=1}^{C}$, computes verifiable rewards and a group baseline (e.g., a rank or mean-normalized signal), and updates $\pi_\theta$ with clipped policy ratios as in PPO \citep{schulman2017proximalpolicyoptimizationalgorithms}. GRPO typically improves \texttt{pass@1} on math/code under RLVR~\citep{shao2024deepseekmath}. However, absent any explicit term for diversity, it can \emph{reduce} variation across attempts as the policy sharpens around locally high-reward regions \citep{dang2025assessing}. 

As some intuition for this distribution change, suppose that the model is attempting to predict the correct answer in a setting where it believes that the answer is \texttt{Yes} with probability 70\% and \texttt{No} with probability 30\%. Cross-entropy loss encourages a model to predict the correct distribution of 70\% \texttt{Yes} and 30\% \texttt{No}; on the other hand, GRPO training would cause the model to collapse its output distribution towards predicting 100\% \texttt{Yes}, as it maximizes the \texttt{pass@1}. Empirically, this can shrink the entropy of the completion distribution and heighten redundancy among attempts, limiting \texttt{pass@k} improvements even as \texttt{pass@1} increases \citep{dang2025assessing}. 

\subsection{Mutual Information and Skill Discovery}
Maximizing mutual information (MI) between latent variables and observed behavior has been a recurring tool for learning structured, controllable representations~\citep{tishby_information_2000,kingma_auto-encoding_2022,stratos_learning_2020}. 

In generative modeling, InfoGAN~\citep{chen2016infoganinterpretablerepresentationlearning} augments GAN training with a variational lower bound on $\mathcal{I}(c;x)$ to make latent codes $c$ predictably control semantic factors (e.g., stroke thickness for MNIST). In variational autoencoders, InfoVAE~\citep{zhao2018infovaeinformationmaximizingvariational} adds an explicit MI term to counteract posterior collapse and preserve informative latents even with expressive decoders.

In sequential decision making, MI has been used to discover diverse, reusable behaviors without external rewards. Early work such as VIC~\citep{gregor_variational_2016} and DIAYN~\citep{eysenbach_diversity_2018} maximizes $\mathcal{I}(s;z)$ or $\mathcal{I}(\tau;z)$, encouraging skills $z$ whose rollouts visit different parts of state or trajectory space and remain identifiable from observations. InfoGAIL~\citep{li2017infogailinterpretableimitationlearning} extends this to imitation learning by maximizing MI between a latent intention and trajectories to capture multi-modal expert behavior. Subsequent methods bias the MI objective toward long-horizon distinctiveness to avoid trivial short-term variation~\citep{sharma_dynamics-aware_2020, hansen_entropic_2021}. Additional related work on MI is available in \autoref{app:related_work_extention}.

Unsupervised skill discovery in RL can be viewed as maximizing the MI between a latent ``skill'' variable and observed trajectories~\citep{gregor_variational_2016, eysenbach_diversity_2018}. Let $z\in\mathcal{Z}$ index a skill and let $\tau$ denote a trajectory. These methods maximize
\begin{equation}
\label{eq:misl-bg}
\max_{\pi}\; \mathcal{I}(\tau; z \mid x) \;=\; \mathbb{E}\Big[\log p_{\pi}(\tau\!\mid\!x,z) - \log p_{\pi}(\tau\!\mid\!x)\Big]
\;=\; \mathcal{H}(\tau\mid x) - \mathcal{H}(\tau\mid x,z).
\end{equation}
This decomposition clarifies the pressure on the policy: \textit{(i)} to increase marginal entropy $\mathcal{H}(\tau\mid x)$ so that trajectories cover more of the solution space; and \textit{(ii)} to decrease conditional entropy $\mathcal{H}(\tau\mid x,z)$ so that each $z$ induces a reproducible, stable mode. The net effect is a set of distinct, consistent behaviors indexed by $z$ that together span diverse solution strategies.

Our setting is closest in spirit to unsupervised skill discovery (e.g., DIAYN, VIC) and to mutual information-based skill learning~\citep{zheng2024can}, but differs in applying these techniques to language models with RLVR training. We develop an approach for maximizing MI tailored to LLM reasoning, and also connect \texttt{pass@k} performance with the mutual information objective.

\subsection{Other techniques for response diversification}
Beyond MI-based training, several other approaches aim to increase output diversity.  

At inference time, decoding-time diversification alters sampling: increasing temperature, switching to nucleus/top-$k$ sampling~\citep{holtzman2020curiouscaseneuraltext, fan2018hierarchicalneuralstorygeneration}, or perturbing prompts~\citep{qiang_prompt_2024}. While simple, these approaches face limitations: \emph{(i)} they often fail to explore qualitatively distinct solution paths~\citep{nguyen_turning_2025, renze_effect_2024}; \emph{(ii)} they require domain-specific tuning~\citep{wiher_decoding_2022}; and \emph{(iii)} they can trade off against correctness and coherence~\citep{nguyen_turning_2025}. Prompt-cycling can inject domain knowledge (e.g., ``try algebra'' vs. ``try geometry''), but it burdens users with prompt engineering and saturates well below human diversity \citep{shur-ofry_growing_2024}.  

Determinantal point processes (DPPs) provide another path by rewarding sets of outputs that span high-volume regions in embedding space~\citep{Kulesza_2012,vijayakumar_diverse_2018, meister_determinantal_2023, wang_diversity_2024}. In reinforcement learning, determinant-based rewards can encourage agents to explore trajectories that span complementary regions of state space~\citep{ash_gone_2021, zhao_efficient_2024}. Compared to MI, which directly couples a latent $z$ with trajectories to ensure reproducible modes, DPP-based diversity is distribution-free: it treats a set of samples as diverse if they occupy a high-volume region in representation space, regardless of whether the same diversity is reproducible under repeated sampling.  

Finally, training-time diversification has also been studied through explicit \texttt{pass@k}-based objectives. \cite{tang2025optimizinglanguagemodelsinference} proposed an unbiased estimator for generic $k$-attempt objectives,
showing overall improved model efficacy. Extending this, \cite{chen2025passktrainingadaptivelybalancing} argue that simply training on \texttt{pass@1} falls victim to a local maximum of over-exploitation and reduced exploration.
They find that \texttt{pass@k} training 
naturally focusing optimization efforts on harder problems
producing significant improvements in both \texttt{pass@k} and \texttt{pass@1}. Outside of verifiable domains,  DivPO~\citep{lanchantin2025diversepreferenceoptimization} alters preference optimization by contrasting diverse high-quality responses with common low-quality ones using a predefined diversity objective, yielding large diversity gains on creative and instruction-following tasks.

As we provide an orthogonal method to improve \texttt{pass@k} and diversity, our approach may complement that of \cite{chen2025passktrainingadaptivelybalancing},  \cite{tang2025optimizinglanguagemodelsinference}, and \cite{lanchantin2025diversepreferenceoptimization}.

\section{Optimizing LLM Diversity with Mutual Information}
\label{sec:method}

\begin{figure}[t]
    \includegraphics[clip, trim=1.7cm 9.5cm 7cm 11cm, width=1.00\textwidth]{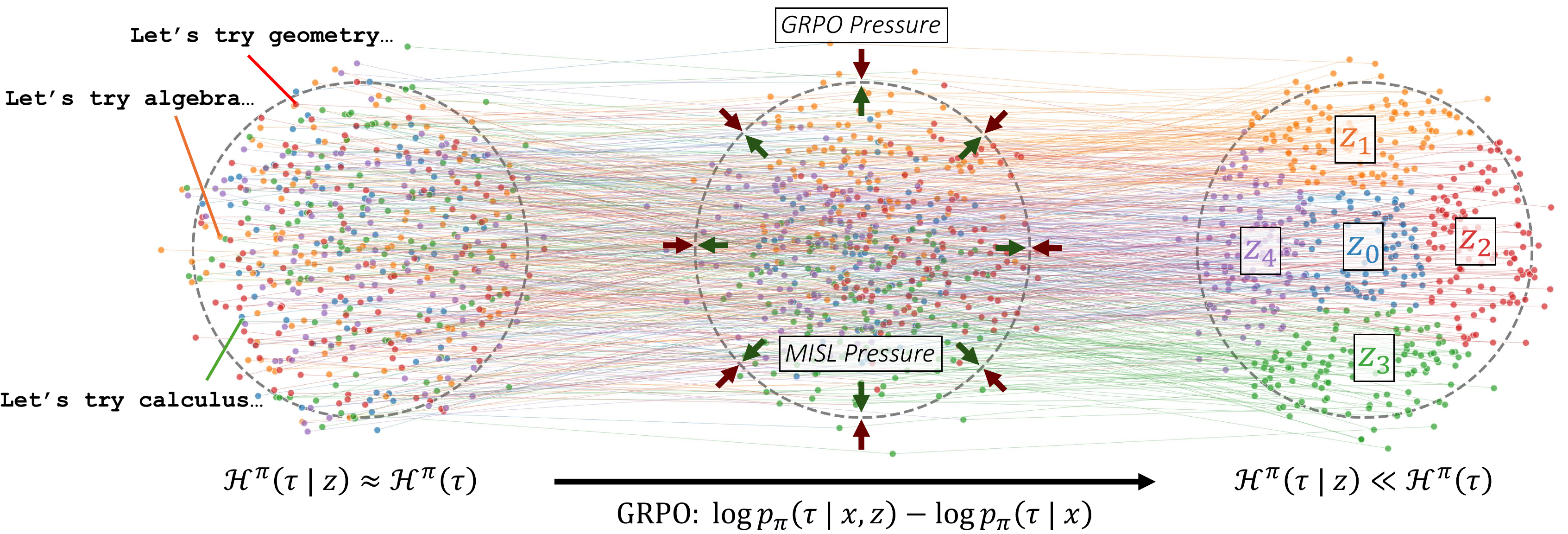}
    
    \caption{Example illustration of how the MISL reward improves \texttt{pass@k} performance. Before MISL (left), the trajectory distribution is independent of the latents $z$, so the conditional entropy is close to the marginal. MISL training prevents distribution collapse due to \texttt{pass@1} training (middle). Adding the token-level MI reward (right) yields well-separated clusters indexed by $z$, reducing conditional entropy while preserving high marginal entropy. At inference, fixing different $z$ values produces consistent and diverse solution strategies.
    \vspace{-5pt}
    \label{fig:overview}
    }
\end{figure}

Given an input $x$ and a policy $\pi(\;\cdot\mid x)$ that produces a completion (trajectory) $\tau=(y_1,\dots,y_T)$, we introduce a \emph{discrete} latent $z\in\{1,\dots,N\}$ via a lightweight prompt prefix (e.g., \texttt{Strategy \{z\} |}), yielding conditional policies $\pi(\cdot\mid x,z)$. During training, $z$ is drawn uniformly at random from the set $\{1,\dots,N\}$. At inference, we select $k\le N$ distinct values of $z$ and generate one completion per value, producing $k$ semantically distinct attempts.

\subsection{Objective}
We would like to encourage \emph{structured response diversity} by maximizing the conditional mutual information $\mathcal{I}(\tau;z\mid x)$. Intuitively, maximizing mutual information makes the outputs of different strategies distinguishable, ensuring that each $z$ induces a reliably different mode. By querying each strategy once, we obtain $k$ semantically distinct attempts. Formally, this corresponds to maximizing
\begin{equation}
\label{eq:misl}
\max_{\pi}\;\mathcal{I}(\tau;z\mid x)
\;=\;\mathbb{E}\Big[\log p_{\pi}(\tau\mid x,z)-\log p_{\pi}(\tau\mid x)\Big],
\end{equation}
which increases the overall entropy of trajectories while reducing the conditional entropy within each $z$-mode, ensuring diverse yet reproducible strategies. The term
\(
p_{\pi}(\tau\mid x)=\frac{1}{N}\sum_{z'=1}^{N} p_{\pi}(\tau\mid x,z')
\)
is a uniform mixture over skills. Maximizing mutual information encourages \textit{(i)} high marginal entropy of trajectories, promoting broad coverage; and \textit{(ii)} low conditional entropy given $z$, so that each response is distinct and determined by $z$. \autoref{fig:overview} provides an overview of the relevant dynamics. 
\subsection{Token-level mutual information reward}
\label{sec:tmi}
We now focus on implementing the mutual information as a token-level reward. For each pair $(x,z)$, let $\{\tau_i\}_{i=1}^{C}$ be $C$ completions sampled from $\pi(\cdot\mid x,z)$. We define a \emph{per-sample} token-level score
\begin{equation}
\label{eq:tmi-per-sample}
r_{\mathrm{TMI}}(\tau_i;x,z)
\;=\;\sum_{t=1}^{|\tau_i|}\Big[\log \frac{p_{\pi}(y_t\mid x,z,y_{<t})}{p_{\pi}(y_t\mid x,y_{<t})}\Big],
\end{equation}
where the second term is the uniform mixture
\begin{equation}
\label{eq:uniform-mixture}
p_{\pi}(y_t\mid x,y_{<t})
\;=\;\frac{1}{N}\sum_{z'=1}^{N} p_{\pi}(y_t\mid x,z',y_{<t}).
\end{equation}
Log-probabilities are computed by $\pi$ on the realized $\tau_i$. In our implementation, the mixture is computed \emph{exactly} across all $N$ skills; this is feasible for the $N$ used in our experiments (Section~\ref{sec:exp-setup}). Since $\frac{1}{C} \sum_i r_{\text{TMI}}(\tau_i;x,z)$ is a Monte Carlo estimator of $\mathcal{I}(\tau;z\mid x)$, we make this our main reward term with \name, with the other reward term being considered in ablation experiments.
Appendix~\ref{sec:smi} discusses an alternative method to calculate this reward based on semantic embeddings rather than the individual tokens.

\subsection{KL regularization}
In standard GRPO training, the per-trajectory KL penalty can be calculated as
\begin{equation}
\Delta_{\mathrm{KL}}(\tau_i)\;=\;\sum_{t=1}^{|\tau_i|}\log\frac{p_\pi(y_t\mid x,y_{<t})}{p_{\pi_{\mathrm{base}}}(y_t\mid x,y_{<t})}.
\end{equation}
While we keep the penalty mathematically the same, the calculation of the numerator needs to be modified to take into account the initial selection of skills before generating a completion. The term in the numerator can be rewritten as in \autoref{eq:uniform-mixture}, with one completion sampled from each skill. As explained in \autoref{sec:eq_deriv}, this regularization term actually incentivizes trained models to approximately satisfy one of our theoretical assumptions.

\begin{figure}[t]
\begin{algorithm}[H]
\caption{\name: A method for training LLMs to produce diverse responses with mutual information.
}
\footnotesize
\label{alg:misl-grpo}
\begin{algorithmic}[1]
\STATE \textbf{Inputs:} base policy $\pi_{\mathrm{base}}$, trainable policy $\pi$, latent count $N$, completions per group $C$, weights $(\alpha_1, \beta)$
\REPEAT
\STATE Sample a minibatch of prompts $\{x\}$
\FOR{each $x$ in the minibatch}
\STATE Sample $z\sim \mathrm{Unif}(\{1,\dots,N\})$; generate $C$ completions $\{\tau_i\}_{i=1}^{C}$ with $\pi(\cdot\mid x,z)$
\STATE Compute $r_{\text{corr}}(\tau_i)$, $r_{\mathrm{TMI}}(\tau_i;x,z)$, and $\Delta_{\mathrm{KL}}(\tau_i)$ as above

\ENDFOR
\STATE Form per-sample rewards via~\autoref{eq:final-reward}; compute advantages; update $\pi$ with GRPO
\UNTIL{convergence}
\end{algorithmic}
\end{algorithm}
\vspace{-2em}
\end{figure}

\subsection{Combined RL objective}
\label{sec:combined}
Let $r_{\text{corr}}(\tau_i)\in\mathbb{R}$ denote the verifiable correctness reward (often binary). The per-sample scalar reward is
\begin{equation}
r(\tau_i;x,z)\;=\;r_{\text{corr}}(\tau_i)\;-\;\beta\,\Delta_{\mathrm{KL}}(\tau_i)
\;+\;\alpha_1\,r_{\mathrm{TMI}}(\tau_i;x,z), \label{eq:final-reward}  %
\end{equation}
with $\alpha_1,\beta\ge 0$. We apply GRPO to optimize the sum of the combined rewards.

\subsection{Training procedure}
\label{sec:algorithm}

We fine-tune a trainable policy $\pi_{\theta}$ with GRPO while injecting a discrete strategy variable $z\in\{1,\dots,N\}$. At each step, we draw a minibatch of prompts $x$ and, for each $x$, sample a strategy $z$ uniformly and generate $C$ completions $\tau_{1:C}\sim\pi_{\theta}(\cdot\mid x,z)$ under fixed decoding. For every completion $\tau$, we compute: \textit{(i)} a verifiable correctness reward $r_{\text{corr}}(\tau)$ from the task’s deterministic checker; \textit{(ii)} the token-level MISL term $r_{\mathrm{TMI}}$ that measures how specific the trajectory is to the chosen strategy;
 and \textit{(iii)} a KL control term toward a frozen base policy.
 We then update the policy with GRPO on this reward. Alg.~\ref{alg:misl-grpo} summarizes our training algorithm.

\subsection{Inference}
\label{sec:inference}
Given a budget of $k$ attempts, we choose $k$ distinct latents from $\{1,\dots,N\}$ and generate one completion per latent under fixed decoding hyperparameters. Aggregation (e.g., majority vote) can optionally be applied. Because each completion is produced by a trained, distinct mode, conditional success probabilities remain larger than with redundant samplings, improving multi-attempt metrics.

\section{Theoretical Connection Between \texttt{pass@k} Improvement and Mutual Information}
\label{sec:main_theory}

Our main theoretical result shows that the mutual information objective is closely tied to \texttt{pass@k}. In particular, we will show that the mutual information objective is a lower bound on \emph{improvement} in the \texttt{pass@k} objective, so maximizing mutual information provably results in an increased (lower bound on) \texttt{pass@k}. Our theoretical results will require the following assumptions:
\begin{enumerate}[itemsep=0pt]
    \item $k$-uniform mixture model: Assume that the marginal distribution over the skills is identical to the base model.
    \item Distributional impact: Let $a_z$ be the probability of success of strategy $z$ and $a$ be the probability of success of the base model. Assume that for all $x\in\mathcal{X}$ there exists $\eta>0$ such that for all $z\in[k]$, $|a_z-a|\geq\eta\delta(\pi_{M,z}(\cdot\mid x),\pi_B(\cdot\mid x))$, where $\delta$ is the total variation distance.
\end{enumerate} 
The second assumption says that the distribution shifts induced by UpSkill correspond to different problem approaches, and, as a result, will have different probabilities of success.
A more precise definition of $k$-uniform mixture models, additional justification for the assumptions, and the statement and proof of the lemma are in \autoref{sec:eq_deriv}. %
\begin{lemma}\label{eq:linearized_improvement} \label{lemma:1}
Let $\texttt{pass@k}_B$ be the \texttt{pass@k} score of the base model on prompt $x$ and $\texttt{pass@k}_M$ be the \texttt{pass@k} score of the mixture model on prompt $x$. Under the above assumptions, we show that:
\begin{align*}
    1-\exp\left(-C_1\eta^2\mathcal{I}(\tau;z\mid x)^2\right) & \leq \frac{\texttt{pass@k}_M-\texttt{pass@k}_B}{1-\texttt{pass@k}_B} \\
    & \leq1-\exp\left(-C_2\mathcal{I}(\tau;z\mid x)\right),
\end{align*}
where $C_1$ depends on $k$ and $C_2$ depends on $k$ and $\max_za_z$.
\end{lemma}

The quantity in the middle can be interpreted as the fraction of possible improvement from the base model that is realized by the mixture model. Since monotonically increasing functions of $\mathcal{I}(\tau;z\mid x)$ provide both lower and upper bounds on how much the mixture model improves over the base model, it makes sense to optimize directly for the mutual information. \name{} explicitly increases $\mathcal{I}(\tau;z\mid x)$ during training, ensuring diversity across
skills and giving a guaranteed improvement in \texttt{pass@k} over the base model.

\section{Experiments}
\label{sec:exp-setup}

We evaluate whether conditioning on a discrete latent $z$ and training with our token-level
mutual-information reward \eqref{eq:tmi-per-sample} improves multi-attempt metrics. We present results in two settings: \textit{(1)} a controlled arithmetic environment that allows for fully verifiable evaluation and
direct inspection of distributional effects, and \textit{(2)} the GSM8K benchmark across three open-weight
models. We then report various experiments that better isolate the impact of \name.

\begin{figure}[t]
  \centering
  \includegraphics[clip, trim=4.1cm 10.3cm 3.9cm 4.8cm, width=0.9\textwidth]{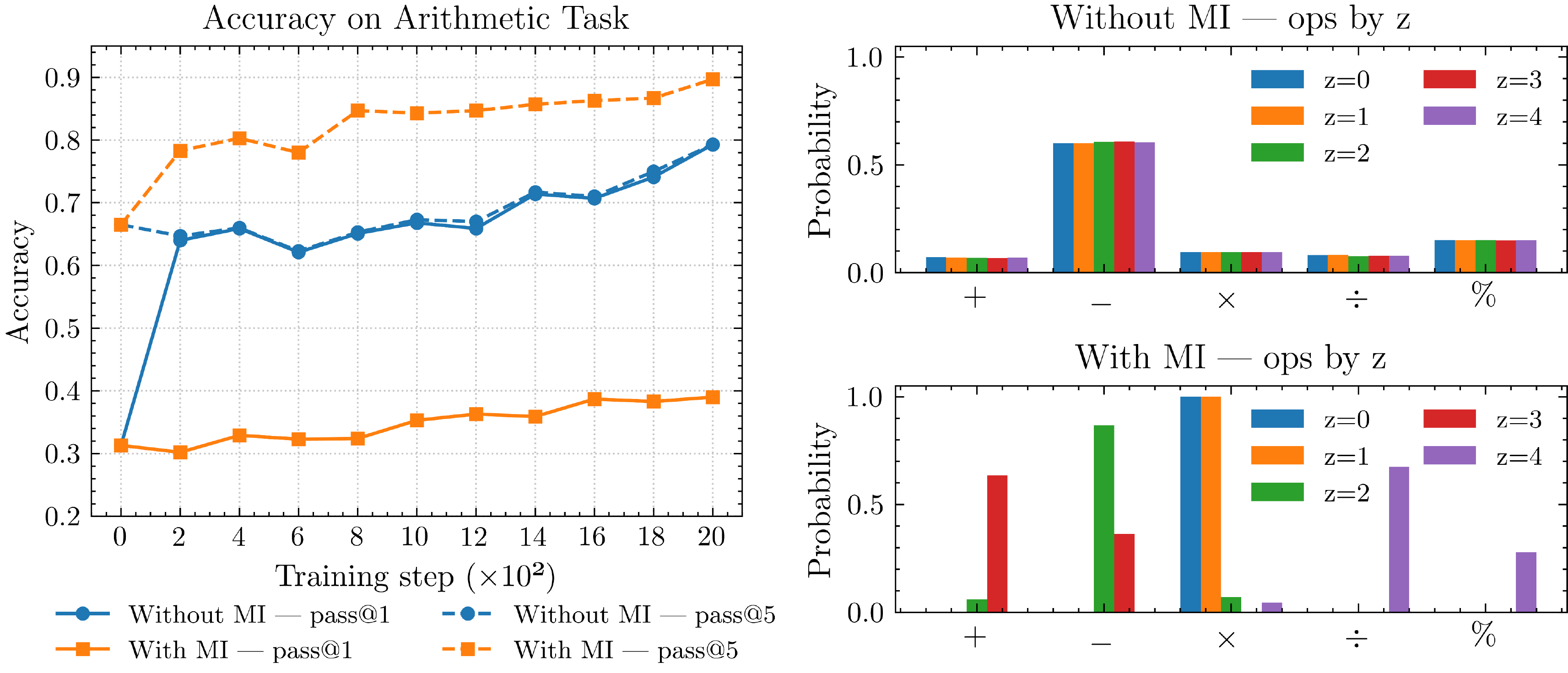}
  
  \caption{Arithmetic environment results. Training curves show that under GRPO alone (blue), \texttt{pass@1} and \texttt{pass@5} converge together, indicating that multiple attempts provide little benefit. With MISL (orange; $N{=}5$), \texttt{pass@5} improves substantially while \texttt{pass@1} remains modest, demonstrating that different latents yield complementary solutions. Operator distributions further highlight this effect: without MISL, they are nearly identical across $z$, reflecting a lack of specialization, whereas with MISL, distinct latents focus on different operators, producing diverse strategies that drive multi-attempt gains.}
  \vspace{-10pt}
  \label{fig:arithmetic}
\end{figure}

\subsection{Arithmetic Environment}
The arithmetic environment consists of prompts with three single-digit integers and a latent skill
index $z\!\in\!\{0,\dots,N{-}1\}$. A small transformer model chooses one of the integers to be a target and is
required to produce a simple arithmetic expression with the other two digits that evaluates to this target. We 
use an automatically verified correctness reward, and training uses GRPO without a KL penalty. Full details of the environment, model, and training procedure are provided in
Appendix~\ref{app:toy-details}.

Figure~\ref{fig:arithmetic} illustrates that in the control condition ($\alpha_1{=}0$), training quickly collapses to a single deterministic strategy: by the end of training, \texttt{pass@1} and \texttt{pass@5} are indistinguishable ($0.793$), offering no benefit from multiple attempts. In contrast, with \name ($\alpha_1{=}0.5$, MI reward cap at $1.0$), the model maintains diverse trajectories, yielding a substantially higher multi-attempt accuracy ($\texttt{pass@5}=0.897$) despite a lower single-attempt accuracy ($\texttt{pass@1}=0.390$). This difference aligns with the entropy dynamics: \name preserves broad output distributions (token entropy changed from $0.723$ to $0.768$ over training), sustaining diverse strategies and higher \texttt{pass@5}. By contrast, the control run—while substantially improving \texttt{pass@1}—collapses to near-deterministic outputs (with entropy $0.723$ changing to  $0.013$), leaving \texttt{pass@5} identical to \texttt{pass@1} (see Appendix~\ref{app:toy-outcomes} for more details).

Figure~\ref{fig:arithmetic}
illustrates that under \name, different $z$ values yield distinct distributions over operators and
digits, whereas the control produces nearly identical distributions across $z$. In this small-scale environment, we can directly observe the learned strategies, and notably $z = 1$ and $z = 2$ converge to risky yet common modes, whereas other values of $z$ cover the remaining operations, improving multi-attempt success with a strategy infeasible for optimizing a \texttt{pass@1} objective. The distributions over the first digit are available in Appendix~\ref{app:digit-distribution}.

We additionally ablate the impact of starting model capabilities, the coefficient of KL penalty, and GRPO parameters (see Appendix \ref{app:toy_with_kl} for full details and results). KL penalty $\beta$ discourages entropy collapse by ensuring the new policy remains close to the initial policy, thereby improving performance. On models with $\beta \in [0.05, 0.10]$, we test $\alpha_1 \in [0.1, 0.3, 0.5]$ and find that well-chosen MI-reward parameters increase $\texttt{pass@k}$ by an average of $3\%$ for the weaker base model. However, for the stronger base model, we find the opposite trend. It is always best to choose $\alpha_1 = 0$, with the best choice of $\alpha_1 \in [0.1, 0.3, 0.5]$ still leading to a $1.2\%$ performance decrease. Our theoretical results in Lemma \ref{eq:linearized_improvement} suggest that \name improvement is negatively related to \texttt{pass@1} (equivalently $\texttt{pass@k}_B$) capability, and thus, although surprising, this result is in line with our theoretical analysis. We separately conjecture that $\beta$, which corresponds with a decrease in exploration from the base policy, conflicts with the mutual information incentive to explore.

\subsection{GSM8K}
\label{sec:gsm}
We next evaluate on GSM8K \citep{cobbe2021training}, a dataset of grade-school arithmetic word
problems. We use 2{,}000 training problems and test on a held-out set of 500 questions. 
All experiments
are conducted in a zero-shot setting with a maximum sequence length of 1024 tokens. We train
LoRA adapters (approximately 80M trainable parameters) on top of three open-weight backbones:
Llama~3.1--8B \citep{grattafiori2024llama3herdmodels}, Qwen~2.5--7B \citep{qwen2025qwen25technicalreport}, and
R1-Distilled--Qwen2.5--Math--1.5B \citep{guo2025deepseek}. We postpone discussion of the R1 results to \autoref{sec:diff_model_families}.

We apply GRPO with a correctness reward and no KL penalty, and \name is applied at the token level as in \autoref{eq:tmi-per-sample} for $N = 5$. We evaluate three learning rates for the \textit{Without MI} baseline ($1\times10^{-4}$, $3\times10^{-5}$, $1\times10^{-5}$). For \textit{With MI} training, we test all combinations of learning rates $\{3\times10^{-5}, 1\times10^{-5}\}$ and $\alpha_{1}\in\{1,5\}$.
We report results from all successful runs, defined by improved correctness during training and mutual information rewards significantly above zero for \textit{With MI} runs. In practice, this is a minimal criterion that helps isolate the effect of the MI reward and excludes runs with degenerate training dynamics. More details, ablations, and a discussion on training stability are available in Appendices \ref{app:gsm8k}, \ref{app:ablations}, and \ref{app:behavioral_diff}.

\begin{figure}[t]
    \centering
    \includegraphics[width=0.9\linewidth]{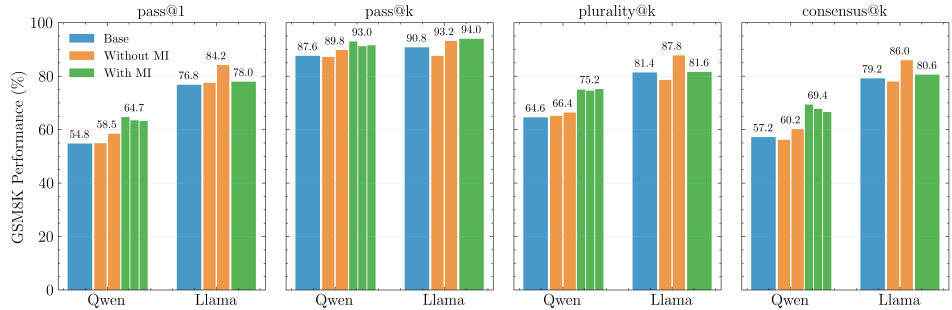}

    \caption{Performance on $500$ held-out problems with $N{=}5$ strategies. We observe gains on all metrics for the Qwen model. \texttt{Base} refers to the model before GRPO training, \texttt{Without MI} refers to after GRPO training without token MI, and \texttt{With MI} refers to training with correctness rewards and token MI. We test multiple configurations for \textit{With MI} and \textit{Without MI} and plot all successful runs as multiple bars, as elaborated in Appendix \ref{app:gsm8k}. For entries with multiple bars, the labeled value is the maximum.}
    \vspace{-5pt}
    \label{fig:z5-perf}
\end{figure}

We include summarized outputs from Qwen in Figure \ref{fig:example_generations} and the isolated Qwen results in Figure \ref{fig:qwen_solo_results}. Figure \ref{fig:z5-perf} showcases performance on the evaluation sets for the Qwen and Llama models. Token-level MISL significantly improves results for Qwen~2.5--7B and Llama~3.1--8B on \texttt{pass@k}.

Interestingly, our results do not entirely parallel the arithmetic environment, as \name{'s} improvement has not come at the cost of \texttt{pass@1}. We conjecture this is because the model's 
final answer is more sensitive to changes in trajectories in the arithmetic environment. If any of the three output tokens in the arithmetic setting are modified, the result of the evaluated expression will almost always be different. In contrast, in the textual reasoning setting, the model can significantly change its reasoning methods without arriving at a different final answer. 
\citet{chen2025passktrainingadaptivelybalancing} have separately found that \texttt{pass@k} training methods can improve \texttt{pass@1} performance, and explicitly mention that their work is supplementary to entropy-guided approaches, such as \name. Finally, although \name's main focus is not on creating interpretable strategies, we provide a discussion of interpretability in \autoref{sec:interpretability}.

\subsection{Ablations and Training Dynamics}
\label{sec:additional_experiments}
\begin{figure}
    \centering
    \includegraphics[width=1\linewidth]{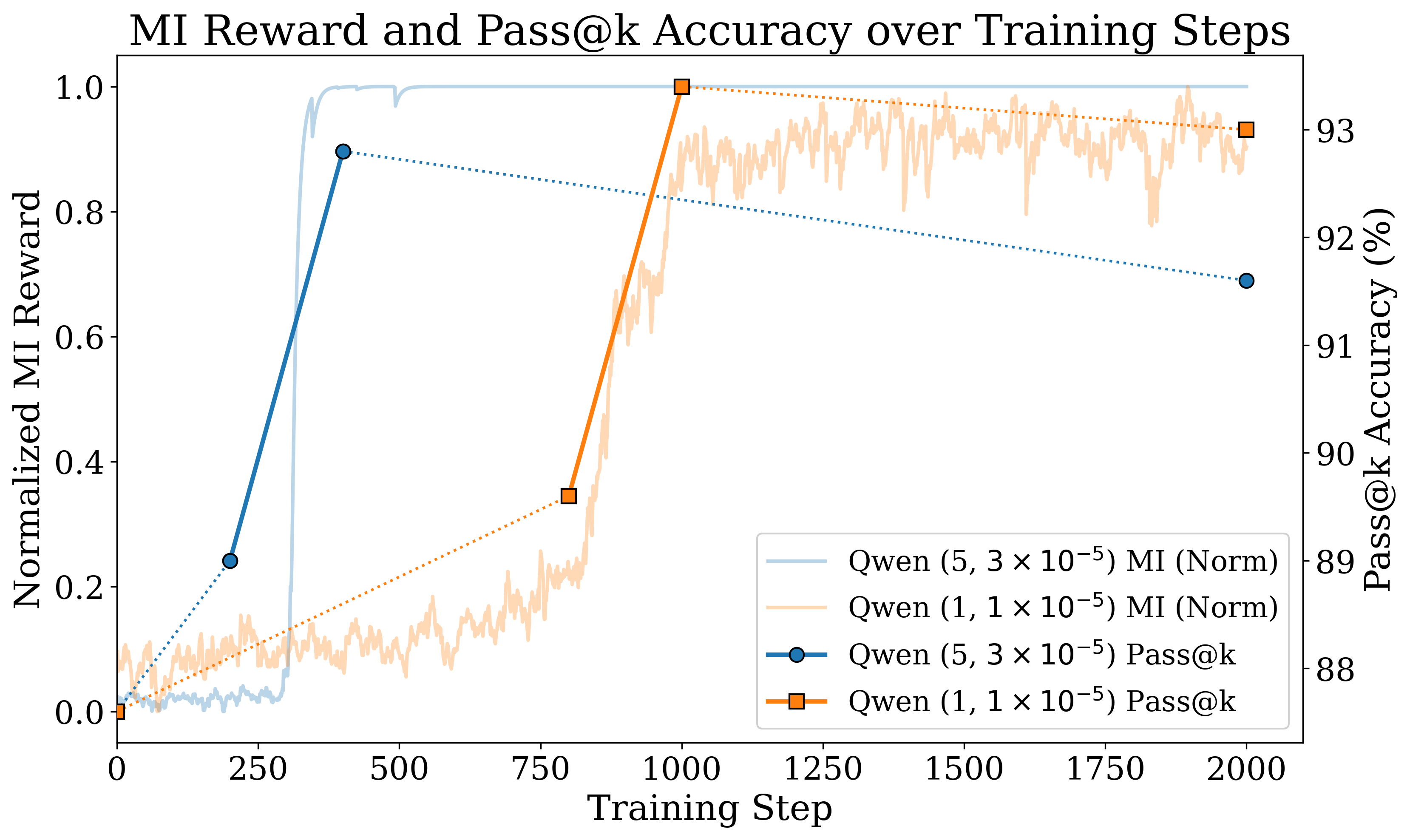}
    \caption{GSM8K pass@5 scores over selected checkpoints of two training processes, overlaid on the mutual information rewards during training. The tuples in the legend are of the form $(\alpha_1, \text{lr})$.}
    \label{fig:passk_evolution}
\end{figure}

We conduct several extensions of the GSM8K experiment to probe the robustness of the approach. First, as we observe from training graphs that the model usually learns to maximize mutual information within a span of 250 training steps, we evaluate
directly before and after this span.
We find that most of the eventual improvement in \texttt{pass@k} happens in this range%
, as shown in \autoref{fig:passk_evolution}.

As further verification, we train models with only the mutual information reward and without the correctness reward. For Qwen and Llama, a high learning rate and early stopping give improvements in \texttt{pass@k} over the base models. 
Repeating this ablation with KL regularization
confirms that \texttt{pass@k} performance can increase without using correctness labels, although the method is most effective when using both MI and correctness rewards, as shown in \autoref{tab:short_results}. This is remarkable as it shows \textbf{ \name can improve \texttt{pass@k} performance without ground-truth correctness labels.}
\begin{table}[ht]
\centering
\small
\begin{tabular}{@{}l l r l r@{}}
\toprule
\textbf{Model} 
& \textbf{Method} 
& \(\boldsymbol{\Delta}\textbf{pass@k}\) 
& \textbf{Method} 
& \(\boldsymbol{\Delta}\textbf{pass@k}\) \\
\midrule
\multirow{2}{*}{Qwen}
& Correctness        & +3.8\% 
& MI                 & +5.2\% \\
& MI + correctness   & +7.0\% 
& MI + KL            & +4.0\% \\
\midrule
\multirow{2}{*}{Llama}
& Correctness        & +2.8\% 
& MI                 & +0.8\% \\
& MI + correctness   & +3.6\% 
& MI + KL            & +2.4\% \\
\bottomrule
\end{tabular}
\vspace{5pt}
\caption{Improvement in \texttt{pass@k} relative to the base model for MI-based methods. Full results can be found in \autoref{tab:combined_results_extended}.}
\vspace{-10pt}
\label{tab:short_results}
\end{table}

Additional GSM8K ablations included increasing testing $N{>}5$ and introducing a semantic MI reward, which we formally introduce in \autoref{sec:smi}. Both of these experiments produced statistically insignificant or negative results.

We also test out-of-distribution capabilities by evaluating models trained on GSM8K on the HumanEval coding benchmark~\citep{chen_evaluating_2021}. Generally, we find that \textit{With MI} runs do not outperform \textit{Without MI}; with the \textit{With MI} Llama run substantially underperforming the base model.

Full details of all experiments are available in Appendix \ref{app:ablations}.

\subsection{Differences between model families}
\label{sec:diff_model_families}
We additionally performed the GSM8K experiment and ablations on the R1-Distilled--Qwen2.5--Math--1.5B model \citep{guo2025deepseek}. However, without additional interventions, we found a decrease in performance across all metrics when training with mutual information and correctness rewards, with \texttt{pass@k} metrics shown in Table \ref{tab:short_results_r1}. We discuss reasons this may occur in Appendix \ref{app:behavioral_diff}. In brief, we hypothesize that R1’s smaller capacity and heavier prior optimization make it more sensitive to the MI objective and to adapter-based fine-tuning.

\begin{table}[ht]
\centering
\small
\begin{tabular}{@{}l l r r l r r@{}}
\toprule
\textbf{Model}
& \textbf{Method}
& \(\boldsymbol{\Delta}\textbf{pass@k}\)
& \textbf{Method}
& \(\boldsymbol{\Delta}\textbf{pass@k}\) \\
\midrule
\multirow{2}{*}{R1}
& Correctness         & +5.2\%
& MI                  & -8.4\% \\
& MI + correctness   & -9.8\%
& MI + KL             & +0.8\% \\
\bottomrule
\end{tabular}
\vspace{5pt}
\caption{Changes in \texttt{pass@k} relative to the base model for the R1 model fine-tuned on correctness only or using our method. Full results can be found in \autoref{tab:combined_results_extended}.}
\vspace{-10pt}
\label{tab:short_results_r1}
\end{table}

\section{Conclusion}

Our experiments show that \name provides a simple and effective way to induce strategy-level diversity in LLMs, leading to gains on multi-attempt metrics such as \texttt{pass@5}. By conditioning on discrete latent variables, the model learns reproducible modes of reasoning that reduce redundancy across attempts and increase the likelihood of success.
\name provides a principled training-time approach for improving response diversity, and our analysis links $\mathcal{I}(\tau;z\mid x)$ to upper bounds in \texttt{pass@k} improvement in training. We hope this stimulates research on robust semantic diversity signals and theoretical ties between information-theoretic objectives and multi-attempt success.
\paragraph{Limitations.}
There are a few notable limitations with our work. Theoretically, although Assumption 1 is possible to softly enforce, it is difficult to strictly enforce, and we hope in future work to reapproach the theoretical guarantees from the perspective of Tsallis entropy \citep{furuichi_information_2006}.
We additionally hope to better understand sources of instability, test \name across more domains and larger models, and better understand the difference in performance of \name across model families.

\paragraph{Reproducibility.}
We make several efforts to encouraging reproducibility of our work and results. \textbf{We have included all experiment code at \href{https://github.com/dshah02/upskill}{https://github.com/dshah02/upskill}}, along with instructions to reproduce our results.
We use only models under permissive licenses. For our theoretical results, we provide a description of assumptions and the complete proof of claims in the appendix. Additionally, for various experiments whose full results could not fit in the main paper, we include the full results in the appendix. 
\section{Acknowledgments}
We would like to especially thank the authors Chongyi Zheng and Benjamin Eysenbach for providing invaluable feedback. We also thank Kevin Wang and Catherine Ji for their helpful discussions and suggestions on promising future directions. The author(s) are pleased to acknowledge that the work reported on in this paper was substantially performed using Princeton University's Research Computing resources. This material is based upon work supported by the National Science Foundation under Award No. 2441665. Any opinions, findings, and conclusions or recommendations expressed in this material are those of the author(s) and do not necessarily reflect the views of the National Science Foundation.

\appendix
\onecolumn
\section{Use of LLMs}
\label{app:use_of_llms}
Large language models were used in the preparation of this work for writing assistance (including polishing, improving presentation of concepts, and restructuring of text), for retrieval and discovery of related work, and for support in producing experimental code and figures. Language models were additionally used for feedback on the paper and to formalize mathematical arguments. All analysis, experimental and method design, and final interpretations are our own. LLM outputs were always rigorously reviewed.

\section{Extended Mutual Information Related Work}
\label{app:related_work_extention}

Estimating MI reliably is challenging in high dimensions. Variational bounds (Barber–Agakov) optimize a classifier or regressor $q_\phi(z\!\mid\!\cdot)$ as a proxy for the intractable posterior~\citep{oord2019representationlearningcontrastivepredictive}. Contrastive bounds such as InfoNCE~\citep{oord2019representationlearningcontrastivepredictive} reduce MI estimation to noise-contrastive classification and have become standard due to their stability. Neural MI estimators like MINE~\citep{belghazi2021minemutualinformationneural} directly optimize a Donsker–Varadhan bound but can suffer from bias/variance trade-offs and training instability. Nonparametric $k$NN estimators (KSG)~\citep{kraskov_estimating_2004} avoid parametric critics but require many samples and are sensitive to dimension, motivating careful batching and normalization when used inside policy gradients. In text generation, MI-style objectives have been used to prevent latent collapse and enable controllable generation, e.g., by encouraging informative latents in variational text models~\citep{zhao2017learningdiscourseleveldiversityneural,zhao2018adversariallyregularizedautoencoders} or aligning codes with style attributes~\citep{john2018disentangledrepresentationlearningnonparallel}. These approaches typically maximize MI between prompts or attributes and latent variables, rather than between a discrete strategy and the full trajectory distribution, and are optimized with supervised losses rather than RL.

Conceptually, our objective reconciles two desiderata emphasized in prior MI work: \emph{coverage} (high marginal entropy over trajectories) and \emph{control} (low conditional entropy given $z$). Whereas decoding-time diversity manipulates token entropy without guarantees about identifiable modes, MI-based diversification learns \emph{reusable, reproducible} modes indexed by a small discrete latent. This makes diversity a first-class, training-time property that can be cleanly exercised at inference by selecting distinct $z$ values.
\section{Statement and derivation of theoretical bounds}
\label{sec:eq_deriv}
\subsection{Problem setup and assumptions}
Let $\mathcal{X}$ be the set of all possible prompts. The statement of \autoref{eq:linearized_improvement} applies to the general class of \emph{$k$-uniform mixture models}.
\begin{definition}
A $k$-uniform mixture model $(M,B)$ is defined to be an ordered pair of a \textit{mixture model}, which is a set of $k$ different policies for generating trajectories, which we will call $\pi_{M,z}(\cdot\mid x)$ for a prompt $x\in\mathcal{X}$ and strategy $z\in[k]$, along with a \textit{base model} $\pi_B(\cdot\mid x)$ for generating trajectories subject to the condition that \[\frac{1}{k}\sum_{z=1}^k\pi_{M,z}(\cdot\mid x)=\pi_B(\cdot\mid x)\,\forall\, x\in\mathcal{X}.\]
\end{definition}
This definition can be interpreted as follows: $\pi_B$ is the trajectory distribution of the original, non-strategy conditioned language model. If we weigh each strategy as being equally important, we sample once from the mixture model by randomly choosing one strategy. In this case, the joint distribution of trajectories from the mixture is \[\pi_M(\cdot\mid x):=\frac{1}{k}\sum_{z=1}^k\pi_{M,z}(\cdot\mid x).\] The condition essentially means that the joint distribution of trajectories over picking a strategy uniformly at random must be the same as the original distribution. Therefore, in essence, the mixture model partitions $\pi_B$ into $k$ different policies that together average back to $\pi_B$.

In practice, strictly enforcing this condition imposes undue constraints on the strategy distribution, but we can still softly enforce this condition with KL regularization. Indeed, writing the KL penalty as $-\beta\Delta_{\mathrm{KL}}(\pi_M\parallel\pi_B)$, we see that maximizing this term corresponds to minimizing the KL distance between $\pi_M$ and $\pi_B$. When training with a nonzero $\beta$, we do see increased training stability while using this method - see \autoref{app:toy_with_kl} for results on the arithmetic environment and \autoref{sec:gsm_kl_results} for results on the GSM8K environment. Also, while in practice one may actually train $N>k$ different strategies and then randomly sample $k$ different strategies so that they still have equal probabilities of being selected, here we make the simplifying assumption that $N=k$, which is true for all of our experiments.

For ease of notation, let $a=\Pr(Y_x(\tau_{B,1})=1\mid x)$ and $a_z=\Pr(Y_x(\tau_{M,z})=1)$ for $z\in[k]$. Because the trajectories $\tau_{M,z}$ are sampled independently, we have that
\begin{equation}\label{eq:avgpass}
\pi_M(\cdot\mid x)=\pi_B(\cdot\mid x)\implies\frac{1}{k}\sum_{z=1}^ka_z=a.
\end{equation}

We provide additional justification for the second assumption made in \autoref{sec:main_theory}. First, if the strategies differ in more than just style and contain meaningful semantic differences, we expect that the difference in success should be proportional to how different these two distributions are. The total variation distance measures this distance in trajectory space, while the constant $\eta$ controls how sensitive the success probabilities are to changes in the distribution shift.
\subsection{Extending \texttt{pass@k} to $k$-uniform mixture models}
We now extend the traditional definition of \texttt{pass@k} to fit the setting of $k$-uniform mixture models to leverage the fact that we now have a natural structure for querying $k$ different strategies by varying $z$. This is notably different from the setting in the consistency assumption, which can be interpreted as querying just one strategy uniformly at random.

For a given prompt $x$ and deterministic verifier $Y_x(\tau)$ outputting $1$ if $\tau$ is a valid output on prompt $x$ and $0$ otherwise, and $k$-uniform mixture model $(M,B)$, define $\texttt{pass@k}_M$ be the probability that querying each of these strategies independently exactly once (see Sec.~\ref{sec:inference}) results in at least one correct answer, and define $\texttt{pass@k}_B$ to be the probability that querying $\pi_B$ independently $k$ times results in at least one correct answer. Writing this out mathematically, for each $z\in[k]$ we sample $\tau_{M,z}\sim\pi_{M,z}(\cdot\mid x
)$; then
\begin{equation}
\label{eq:passkm}
\begin{split}
\texttt{pass@k}_M \;=\; 1 - \Pr\!\left(\bigcap_{z=1}^{k}\{Y_x(\tau_{M,z})=0\}\,\middle|\,x\right)=1-\prod_{z=1}^k\Pr(Y_x(\tau_{M,z})=0\mid x).
\end{split}
\end{equation}
While we use the same definition of \texttt{pass@k} for $B$ as in standard literature, we include it for the sake of completeness; similarly, when sampling $\tau_{B,z}\sim\pi_B(\cdot\mid x)$ independently for each $z\in[k]$, we find that
\begin{equation}
\label{eq:passkb}
\begin{split}
\texttt{pass@k}_B \;=\; 1 - \Pr\!\left(\bigcap_{z=1}^{k}\{Y_x(\tau_{B,z})=0\}\,\middle|\,x\right)=1-\Pr(Y_x(\tau_{B,1})=0\mid x)^k
\end{split}
\end{equation} where we simplify the product into the RHS of \eqref{eq:passkb} with the independence of samples.
Then \[\texttt{pass@k}_B=1-(1-a)^k,\texttt{pass@k}_M=1-\prod_{z=1}^k(1-a_z).\]

We now give the precise statement of the main result.
\begin{lemma}[\texttt{pass@k} Improvement for $k$-uniform Mixture Models, Full Statement]
Let $u=\max_{z\in[k]}a_z$ and let $\varphi:\mathbb{R}_{\geq 0}\mapsto\mathbb{R}$ be defined as $\varphi(x)=\frac{x\log x}{x-1}$ for $x\neq 0,1$ and $\varphi(0)=0,\varphi(1)=1$. Then \[1-\exp\left(-\frac{k\eta^2\mathcal{I}(\tau;z\mid x)^2}{2\varphi(k)^2}\right)\leq\frac{\texttt{pass@k}_M-\texttt{pass@k}_B}{1-\texttt{pass@k}_B}\leq1-\exp\left(-\frac{k\mathcal{I}(\tau;z\mid x)}{4(1-u)^2}\right).\]
\end{lemma}
\subsection{Derivation of Lower Bound}
\label{sec:lowerbound}
Using Taylor's Theorem on $f(y)=\log(1-y)$ gives the equations $f(a_z)=f(a)+(a_z-a)f'(a)+\frac{1}{2}(a_z-a)^2f''(\xi_z)$ where $\xi_z$ lies in between $a_z$ and $a$, for $z\in[k]$. Summing all of these equations, the linear terms cancel due to \eqref{eq:avgpass}. Then \begin{equation}\label{eq:taylorexp}
\sum_{z\in[k]}\log(1-a_z)=k\log(1-a)+\sum_{z\in[k]}\frac{1}{2}(a_z-a)^2f''(\xi_z).
\end{equation}
Let $A$ be the random variable that takes value $a_z$ for each $z\in[k]$ with probability $\frac{1}{k}$. Since $f''(y)=-\frac{1}{(1-y)^2}$ is a decreasing function in the interval $[0,1)$, we find that $f''(\xi_z)\leq f''(0)=-1$ for all $z\in[k]$, i.e. \[\sum_{z\in[k]}\log(1-a_z)\leq k\log(1-a)-\sum_{z\in[k]}\frac{1}{2}(a_z-a)^2\]\[\implies k\log(1-a)-\sum_{z\in[k]}\log(1-a_z)\geq\frac{k\Var(A)}{2}\]\[\implies\frac{1-\texttt{pass@k}_B}{1-\texttt{pass@k}_M}\geq\exp\left(\frac{k}{2}\Var(A)\right)\]\[\implies\texttt{pass@k}_M-\texttt{pass@k}_B=(1-\texttt{pass@k}_B)\left(1-\frac{1-\texttt{pass@k}_M}{1-\texttt{pass@k}_B}\right)\]\[\geq(1-\texttt{pass@k}_B)\left(1-\exp\left(-\frac{k\Var(A)}{2}\right)\right).\]
We now find a lower bound for $\Var(A)$ in terms of the mutual information to finish off the proof. Let $\delta(\cdot,\cdot)$ represent the total variation distance between two distributions. We have that $\Var(A)\geq\mathbb{E}[|A-a|]^2$ by Jensen's Inequality, so using the distributional impact assumption, \[\Var(A)\geq\left(\mathbb{E}_{z\sim\text{Unif}\{1,2,\ldots,k\}}[|a_z-a|]\right)^2=\left(\frac{1}{k}\sum_{z=1}^k|a_z-a|\right)^2\geq\left(\frac{1}{k}\sum_{z=1}^k\eta\delta(\pi_{M,z},\pi_B)\right)^2\]\[=\eta^2\left(\mathbb{E}_{z\sim\text{Unif}\{1,2,\ldots,k\}}[\delta(\pi_{M,z},\pi_B)]\right)^2.\] Note that by the assumption that $\pi_M(\cdot\mid x)=\pi_B(\cdot\mid x)$ we have that $0\leq\pi_{M,z}(\cdot\mid x)\leq k\pi_B(\cdot\mid x)$ for all $z\in[k]$. Therefore, \[\frac{d\pi_{M,z}}{d\pi_B}(\tau)\in[0,k].\] Applying Theorem 26 from \citet{sason_f-divergence_2016} with $\beta_2=0,\beta_1=\frac{1}{k}$ where the constants are from the assumption on bounded likelihood ratio, we have that $D_{\text{KL}}(\pi_{M,z}\parallel\pi_B)\leq\varphi(k)\delta(\pi_{M,z},\pi_B)$. Summing over $z\in[k]$ and dividing by $k$ yields \[\mathcal{I}(\tau;z\mid x)=\mathbb{E}_{z\sim\text{Unif}\{1,2,\ldots,k\}}[D_{\text{KL}}(\pi_{M,z}\parallel\pi_B)]\leq\varphi(k)\mathbb{E}_{z\sim\text{Unif}\{1,2,\ldots,k\}}[\delta(\pi_{M,z},\pi_B)].\] Putting both bounds together, we finally find that \[\Var(A)\geq\left(\frac{\eta\mathcal{I}(\tau;z\mid x)}{\varphi(k)}\right)^2\implies\texttt{pass@k}_M-\texttt{pass@k}_B\geq(1-\texttt{pass@k}_B)\left(1-\exp\left(-\frac{k\eta^2\mathcal{I}(\tau;z\mid x)^2}{2\varphi(k)^2}\right)\right)\] as desired.
\subsection{Derivation of Upper Bound}
Let $u=\max_za_z<1$. In the interval  $[\min(a,a_1,a_2,\ldots,a_k),\max(a,a_1,a_2,\ldots,a_k)]$ $f''$ achieves its minimum at $f''(u)=-\frac{1}{(1-u)^2}$. Then from \eqref{eq:taylorexp} \[\sum_{z\in[k]}\log(1-a_z)\geq k\log(1-a)-\frac{1}{2(1-u)^2}\sum_{z\in[k]}(a_z-a)^2\]\[\implies \prod_{z=1}^k(1-a_z)\geq(1-a)^k\exp\left(-\frac{1}{2(1-u)^2}\sum_{z\in[k]}(a_z-a)^2\right)\]
\begin{equation}\label{eq:passbound}
\begin{split}
\implies 1-\texttt{pass@k}_M\geq(1-\texttt{pass@k}_B)\exp\left(-\frac{1}{2(1-u)^2}\sum_{z\in[k]}(a_z-a)^2\right).
\end{split}
\end{equation}
This places an upper bound on how much we can possibly improve \texttt{pass@k} compared to our original trajectory distributions. Using Pinsker's Inequality, we find that for all $i\in[k]$, \[|a_i-a|=|\Pr[Y(\tau_{M,i})=1]-\Pr[Y(\tau_{B,i})=1]|=\left|\sum_{Y(\tau')=1}\pi_{M,i}(\tau')-\sum_{Y(\tau')=1}\pi_{B,i}(\tau')\right|\]\[\leq\sum_{Y(\tau')=1}|\pi_{M,i}(\tau')-\pi_{B,i}(\tau')|\leq\sum_{\tau'}|\pi_{M,i}(\tau')-\pi_{B,i}(\tau')|\leq\delta(\pi_{B,i},\pi_{M,i})\leq\sqrt{\frac{1}{2}D_{\text{KL}}(\pi_{M,i}\parallel\pi_{B,i})}\]\[\implies (a_i-a)^2\leq\frac{1}{2}D_{\text{KL}}(\pi_{M,i}\parallel\pi_{B,i})\]\[\implies\sum_{z\in[k]}(a_z-a)^2\leq\frac{k}{2}\cdot\frac{1}{k}\sum_{z\in[k]}D_{\text{KL}}(\pi_{M,i}\parallel\pi_{B,i})\]\[=\frac{k}{2}\mathbb{E}_{z\sim\text{Unif}\{1,2,\ldots,k\}}[D_{\text{KL}}(\pi_{M,i}\parallel\pi_B)]=\frac{k}{2}\mathcal{I}(\tau;z\mid x).\]  Combining this with \eqref{eq:passbound} yields
\begin{equation}\label{eq:main_highlight}
\begin{split}
1-\texttt{pass@k}_M\geq(1-\texttt{pass@k}_B)\exp\left(-\frac{k}{4(1-u)^2}\mathcal{I}(\tau;z\mid x)\right).
\end{split}
\end{equation}
As a result, if $\mathcal{I}(\tau;z\mid x)$ is too small, then our theoretical upper bound on improvement in \texttt{pass@k} between steps $0$ and $T$ will also be very small.

Rearranging \eqref{eq:main_highlight} to bound the improvement \(\Delta := \texttt{pass@k}_M-\texttt{pass@k}_B\), we obtain \[\Delta=(1-\texttt{pass@k}_B)-(1-\texttt{pass@k}_M)\leq(1-\texttt{pass@k}_B)\left(1-\exp\left(-\frac{k}{4(1-u)^2}\mathcal{I}(\tau;z\mid x)\right)\right)\]\[\leq(1-\texttt{pass@k}_B)\cdot\frac{k}{4(1-u)^2}\cdot\mathcal{I}(\tau;z\mid x),\]
where the final inequality uses $1-e^{-x}\le x$.

\section{Semantic mutual information reward}
\label{sec:smi}
One observation that we made empirically is that token-level differences tend to reflect formatting or paraphrasing, rather than semantically distinct strategies. To bias toward more meaningful differences, we test an alternative method for measuring mutual information by embedding completions with a fixed encoder $\psi(\tau)\in\mathbb{R}^{d}$ and estimating the mutual information between embeddings and skills for a \emph{single prompt} $x$:
\begin{equation}
\label{eq:smi-group}
\widehat{\mathcal{I}}\big(\psi(\tau);z\,\big|\,x\big)
\end{equation}
using the Kraskov–Stögbauer–Grassberger (KSG) $k$-nearest-neighbor estimator~\citep{kraskov_estimating_2004}, implemented with the library NPEET~\citep{steeg_gregversteegnpeet_2025}. Concretely, for each $x$ we collect the set of embeddings across strategies and samples,
\(
\mathcal{B}(x)=\{(\psi(\tau_{i}^{(z)}),\,z)\,:\,z\in\{1,\dots,N\},\, i=1,\dots,C\},
\)
and apply KSG to $\mathcal{B}(x)$ to obtain a single scalar $r_{\mathrm{SMI}}(x)$.

\section{Arithmetic Environment}
\label{app:toy-details}

\subsection{Task}
Each problem instance consists of three integers $a,b,c\in\{0,\dots,9\}$. A small transformer model chooses one of the integers to be a target and is
required to produce a simple arithmetic expression with the other two digits that evaluates to this target. Valid operators are $\{+,-,\times,\div,\bmod\}$, with division and modulo defined only when results are integers and denominators are nonzero. A latent skill index $z\in\{0,\dots,N{-}1\}$ is provided as part of the prompt, conditioning the model on which strategy to adopt.

\subsection{Prompt format and conditioning}
The input is formatted as
\[
\texttt{[z] a b c}
\]
where \texttt{[z]} encodes the latent skill id and $a,b,c$ are the three digits. The model is required to generate exactly three tokens in the order \texttt{(digit, operator, digit)}. This restriction enforces that every completion corresponds to a candidate arithmetic expression of the form $L\,o\,R$ with $L,R\in\{a,b,c\}$. The verifier deterministically evaluates the completion, awarding a reward of 1 if the output matches the designated target and 0 otherwise.

\subsection{Evaluation protocol}
At inference, we fix $k=\min(N,5)$ distinct latent skills and generate one completion per skill at a fixed temperature. We then report:
\begin{itemize}[leftmargin=1.5em, itemsep=2pt]
  \item \texttt{pass@1}: the fraction of skills (out of $k$) that yield a correct completion, i.e.\ the probability that a single uniformly sampled skill would succeed.
  \item \texttt{pass@k}: the probability that at least one of the $k$ skills yields a correct completion.
\end{itemize}
This definition differs from conventional \texttt{pass@1} (best-of-$k$) to more closely capture the multi-skill sampling process we target.

\subsection{Model and optimization}
The policy is a 2-layer causal Transformer (hidden size 128, 4 attention heads, pre-layer normalization, GELU activations). Inputs are embedded with learned token and positional embeddings. The output vocabulary consists of 15 symbols: \texttt{0–9}, \texttt{+}, \texttt{-}, \texttt{*}, \texttt{/}, \texttt{\%}. Training uses GRPO updates without a KL penalty, comparing runs with and without the MISL reward. Teacher-forced cross-entropy warmup is applied for 100 steps before switching to RL. Unless otherwise noted: $N{=}5$, temperature $0.9$, batch size 32 groups, and $C{=}5$ completions per update. 
\subsection{Training Outcomes}
\label{app:toy-outcomes}

Table~\ref{tab:toy-outcomes} reports \texttt{pass@1}, \texttt{pass@5}, and marginal token entropy
$H_m$ at the end of the supervised warmup (step 0), mid-training (step 1000), and the final step
(step 2000). All runs use $N=5$ skills and 2000 training steps.

\begin{table*}[ht]
\centering
\begin{tabular}{lccccccccc}
\toprule
& \multicolumn{3}{c}{After Warmup (Step 0)} & \multicolumn{3}{c}{Step 1000} & \multicolumn{3}{c}{Step 2000} \\
\cmidrule(lr){2-4}\cmidrule(lr){5-7}\cmidrule(lr){8-10}
Condition & \texttt{p@1} & \texttt{p@5} & $H_m$ & \texttt{p@1} & \texttt{p@5} & $H_m$ & \texttt{p@1} & \texttt{p@5} & $H_m$ \\
\midrule
Control -- ($\alpha_1=0$) 
& 0.313 & 0.665 & 0.723 
& \textbf{0.668} & 0.673 & 0.016 
& \textbf{0.793} & 0.793 & 0.013 \\
($\alpha_1=0.5$, cap=1.0) 
& 0.313 & 0.665 & 0.723 
& 0.353 & \textbf{0.843 }& 0.755 
& 0.390 & \textbf{0.897} & 0.768 \\
($\alpha_1=0.5$, cap=1.5) 
& 0.313 & 0.665 & 0.723 
& 0.338 & 0.833 & 0.764 
& 0.399 & 0.830 & \textbf{0.852} \\
($\alpha_1=1.0$, cap=1.0) 
& 0.313 & 0.665 & 0.723 
& 0.281 & 0.813 & \textbf{0.813} 
& 0.373 & \textbf{0.897 }& 0.844 \\
\bottomrule
\end{tabular}
\vspace{5pt}
\caption{Arithmetic environment training outcomes. Accuracy is reported as \texttt{pass@1} and
\texttt{pass@5}; $H_m$ is marginal token entropy. MISL prevents entropy collapse and sustains
diverse skill-conditioned strategies.}
\label{tab:toy-outcomes}
\end{table*}

\subsection{Sensitivity to \texorpdfstring{$\alpha_1$}{alpha1} and \texttt{cap}}
\label{app:toy-abl-cap-alpha}

Increasing the clipping parameter \texttt{cap} sustains higher entropy but also introduces more variance across runs. Raising $\alpha_1$ strengthens specialization and increases \texttt{pass@5}, though at the expense of \texttt{pass@1}. A moderate setting of $\alpha_1{=}0.5$ and \texttt{cap}$=1.0$ provided the most consistent balance in our experiments.

\subsection{Additional Distribution Data}
\label{app:digit-distribution}
\begin{figure}[H]
    \centering
    \includegraphics[width=\linewidth]{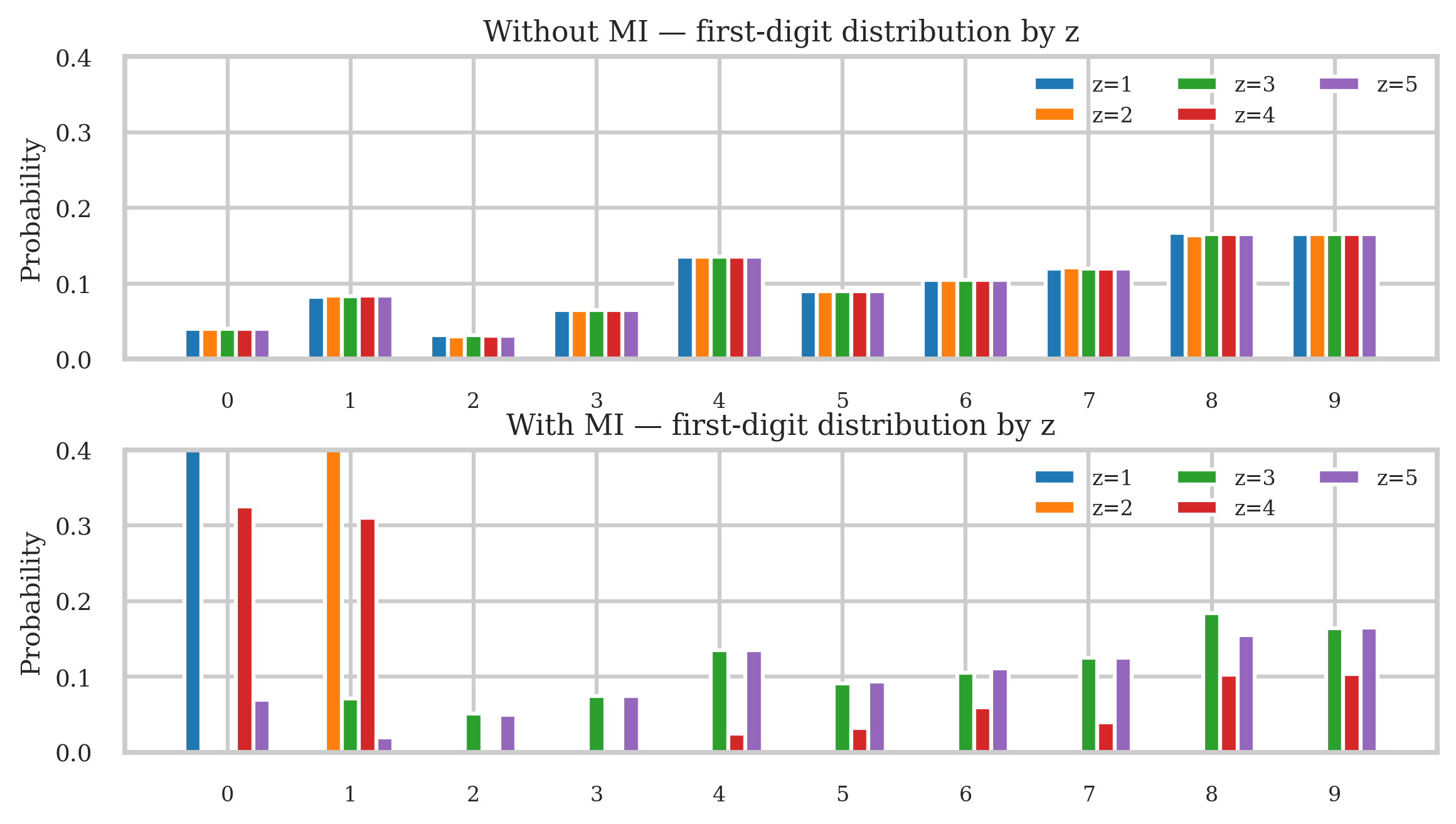}
    \caption{Learned distribution over first response digit with $\alpha_1 = 0.5$ and $\text{cap}=1.0$}
\end{figure}

\section{Ablated Arithmetic Environment}
\label{app:toy_with_kl}

Here we ablate \emph{model capacity} by varying the number of warmup steps and by adding a KL penalty.  
Warmup 50 corresponds to a weaker base model, while warmup 100 produces a stronger base model.  
This manipulation allows us to study how \name interacts with correctness-oriented pretraining and how much headroom remains for diversity improvements.  
We also include a KL penalty with coefficients \(\texttt{kl\_coef}\in\{0.05,0.10\}\), completions per group \(C\in\{5,10\}\), and MIs weights \(\alpha_1\in\{0.0,0.1,0.3,0.5\}\) (token MI only).

\begin{table}[ht]
\centering
\small
\begin{tabular}{@{}lcc@{}}
\toprule
\textbf{Warmup Steps} & \(\textbf{pass@1}\) & \(\textbf{pass@k}\) \\
\midrule
50  & 0.235 & 0.540 \\
100 & 0.313 & 0.665 \\
\bottomrule
\end{tabular}
\vspace{5pt}
\caption{Performance of the base model after warmup only (no RL). Warmup 50 yields a weaker base capacity, while warmup 100 yields a stronger base capacity.}
\label{tab:warmup_capacity}
\end{table}

Tables~\ref{tab:kl_pass1} and \ref{tab:kl_pass5} report \texttt{pass@1} and \texttt{pass@5} after RL across settings. Columns \(\texttt{aX.Y}\) denote \(\alpha_1=X.Y\).

\begin{table*}[ht]
\centering
\small
\begin{tabular}{@{}rrrrrrrr@{}}
\toprule
\textbf{kl\_coef} & \textbf{warmup} & \textbf{C} &
\(\mathbf{pass@1_{a0.0}}\) & \(\mathbf{pass@1_{a0.1}}\) & \(\mathbf{pass@1_{a0.3}}\) & \(\mathbf{pass@1_{a0.5}}\) \\
\midrule
0.050 & 50  &  5 & 0.779 & 0.780 & 0.561 & 0.457 \\
0.050 & 50  & 10 & 0.845 & 0.817 & 0.655 & 0.439 \\
0.100 & 50  &  5 & 0.813 & 0.801 & 0.557 & 0.366 \\
0.100 & 50  & 10 & 0.833 & 0.860 & 0.629 & 0.437 \\
\midrule
0.050 & 100 &  5 & 0.908 & 0.897 & 0.749 & 0.521 \\
0.050 & 100 & 10 & 0.914 & 0.897 & 0.767 & 0.521 \\
0.100 & 100 &  5 & 0.911 & 0.895 & 0.788 & 0.568 \\
0.100 & 100 & 10 & 0.917 & 0.901 & 0.844 & 0.614 \\
\bottomrule
\end{tabular}
\vspace{5pt}
\caption{\texttt{pass@1} after RL with KL penalty in the arithmetic environment.}
\label{tab:kl_pass1}
\end{table*}

\begin{table*}[ht]
\centering
\small
\begin{tabular}{@{}rrrrrrrr@{}}
\toprule
\textbf{kl\_coef} & \textbf{warmup} & \textbf{C} &
\(\mathbf{pass@5_{a0.0}}\) & \(\mathbf{pass@5_{a0.1}}\) & \(\mathbf{pass@5_{a0.3}}\) & \(\mathbf{pass@5_{a0.5}}\) \\
\midrule
0.050 & 50  &  5 & 0.793 & 0.807 & 0.840 & \textbf{0.870} \\
0.050 & 50  & 10 & \textbf{0.867} & 0.833 & \textbf{0.867 }& 0.817 \\
0.100 & 50  &  5 & 0.840 & 0.843 & 0.840 & \textbf{0.847} \\
0.100 & 50  & 10 & 0.857 & \textbf{0.893} & 0.850 & 0.820 \\
\midrule
0.050 & 100 &  5 & \textbf{0.940} & 0.917 & 0.863 & 0.883 \\
0.050 & 100 & 10 & \textbf{0.927} & 0.903 & 0.903 & 0.887 \\
0.100 & 100 &  5 & \textbf{0.927 }& 0.907 & 0.910 & 0.917 \\
0.100 & 100 & 10 & \textbf{0.944} & 0.931 & 0.940 & 0.923 \\
\bottomrule
\end{tabular}
\vspace{5pt}
\caption{\texttt{pass@5} after RL with KL penalty in the arithmetic environment.}
\label{tab:kl_pass5}
\end{table*}
Overall, a modest KL penalty (\(0.05\text{--}0.10\)) prevents entropy collapse and supports higher multi-attempt accuracy, especially when the warmup baseline is stronger (100 vs.\ 50). With warmup 50, larger \(\alpha_1\) increases \texttt{pass@5} substantially (e.g., \(0.793 \rightarrow 0.870\)), though often at the expense of \texttt{pass@1}. With warmup 100, the base capacity is already high, and further MISL gains are more limited, reflecting our theoretical expectation that improvements in \texttt{pass@k}\ depend on the available headroom for diversity.

\section{GSM8K Experimental Details}
\label{app:gsm8k}

\subsection{Setup}
We evaluate our method on GSM8K \citep{cobbe2021training}, a grade-school arithmetic dataset with 2,000 training and 500 held-out evaluation problems. Prompts are provided in a zero-shot format with a maximum sequence length of 1024 tokens. For inference, we fix $k$ distinct latent skill identifiers and sample one completion per skill at fixed temperature.

\subsection{Models}
We attach LoRA adapters (about 80M parameters) to three open-weight backbones: Llama~3.1–8B, Qwen~2.5–7B, and R1-Distilled–Qwen2.5–Math–1.5B. Training uses GRPO with correctness reward only (control) or both correctness and token-level MISL (experimental) varying $\alpha_1$ and the learning rate. The two possible learning rates tested were $1\times 10^{-5}$ and $3\times 10^{-5}$. For control models, we also included a learning rate of $1\times 10^{-4}$. The two possible values of $\alpha_1$ were $1$ and $5$. Thus, for experimental runs, there are 4 configurations and for the control run, there are 3 configurations. Each experiment is run for 2000 steps on a single H100 GPU with 80 GB of memory. For the results in Figure \ref{fig:z5-perf} and in Table \ref{tab:all_runs_detailed}, we include all runs meeting the following criteria: the correctness reward increases over the course of training and, for the experimental runs, the mutual information reward increases from $0$.%

\subsection{Evaluation Details}

We train and evaluate the model with the prompting format of: "\texttt{Strategy [z] | Question}" when training with the token MI reward and with only the question otherwise (base model and without MI models). 
During training and evaluation, we fix the system prompt as ``You are a helpful math assistant that solves problems step by step."

In inference, we fix $N$ distinct latent skills and generate one completion per skill, with $N = 5$ except in ablations. To determine correctness on GSM8K, we use a robust extraction function that searches for values within common tags (such as "\texttt{<answer> </answer>}"), defaulting to the final word containing digits in the model's output, removes characters not in \texttt{0123456789.-}, and compares the resulting value against the reference answer. We then report:
\begin{itemize}
    \item \texttt{pass@1}: the fraction of skills (out of $N$) that yield a correct completion; i.e., the probability that a single uniformly sampled skill would succeed
    \item \texttt{pass@k}: the probability that at least one of the $k$ skills yields a correct completion
    \item \texttt{plurality@k}: the probability that there is a unique mode response, and that it is correct
    \item \texttt{consensus@k}: the probability that a strict majority of the completions are correct.
\end{itemize}

For evaluating models on the HumanEval coding benchmark, we use the system prompt ``You are a helpful assistant who writes correct, well-tested Python code. Generate only the Python code to complete the function described in the docstring. Do not include any explanations, introductory text, or natural language comments outside the function body.'' We use the same prompting format as in the GSM8K case, and for evaluation we first strip away any coding tags like \texttt{```python}, then use the evaluation harness provided by the HumanEval library to evaluate the correctness of the generated code. For this evaluation, we only measure the metrics \texttt{pass@1} and \texttt{pass@5}.
\subsection{Addressing Performance Collapses}
\label{sec:perf_collapse}
\begin{figure}[h]
    \centering
    \includegraphics[width=1\linewidth]{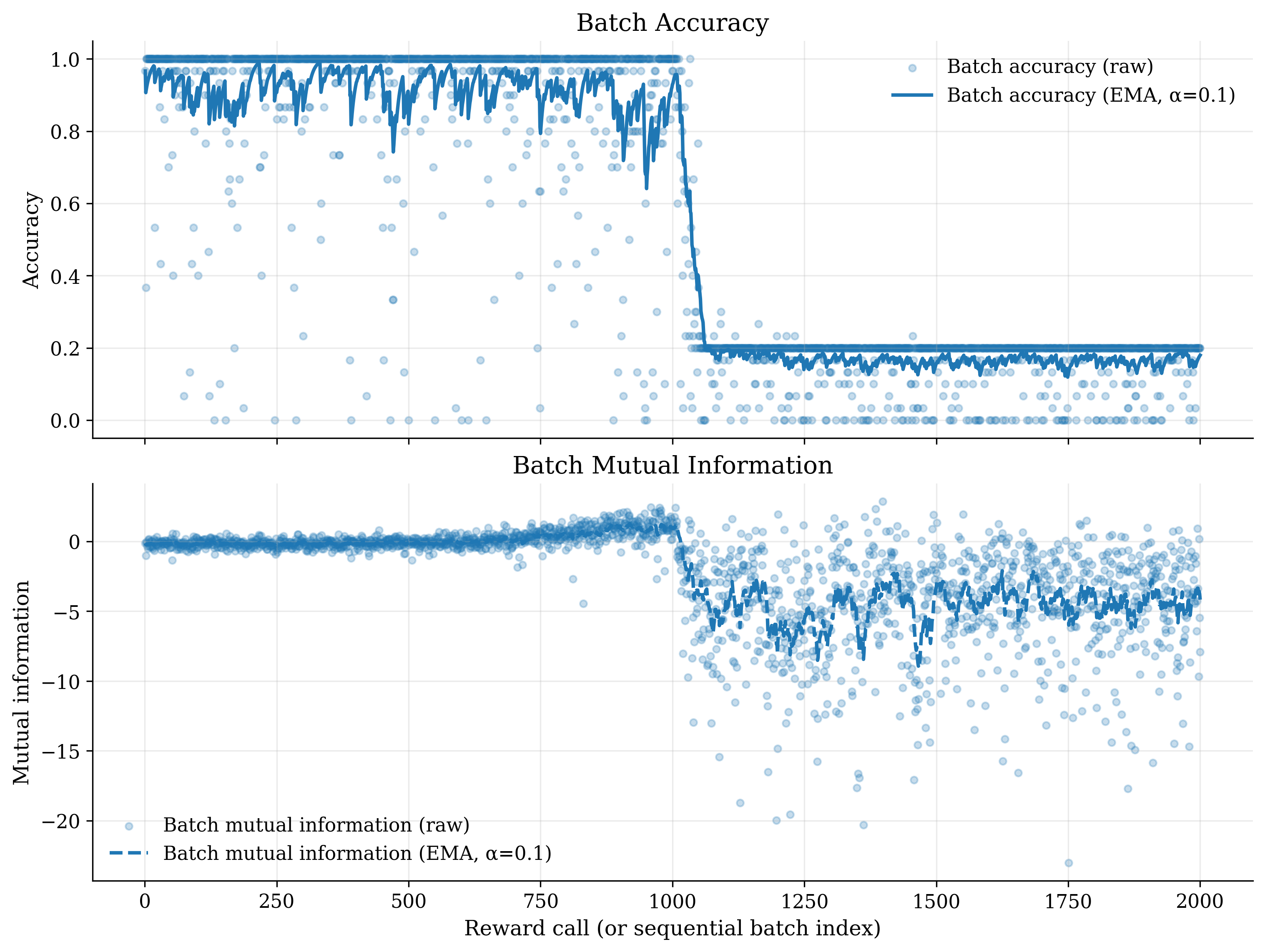}
    \caption{Graph of model rewards during training in which it suffers strategy collapse.}
    \label{fig:modal_collapse}
\end{figure}

Due to compute constraints, we did not evaluate model instances where the training either showed clear signs of performance collapse, which affected both baseline results and some experimental results, or no significant increase in MI reward over the training process. All evaluation results, including hyperparameter choices and both GSM8K and HumanEval results are shown in \autoref{tab:all_runs_detailed}. We found that in the experimental setting, a common reason for performance collapse is the model maximizing mutual information with disjoint ``gibberish" modes. This drastically reduces correctness and is challenging to recover from. An example of reward curves over training on such a strategy collapse can be found in \autoref{fig:modal_collapse}.
\begin{table*}[t]
\centering
\small
\begin{tabular}{@{}lllcrrrrrr@{}}
\toprule
& & & & \multicolumn{4}{c}{\textbf{GSM8K (In-Distribution)}} & \multicolumn{2}{c}{\textbf{HumanEval}} \\
\cmidrule(lr){5-8} \cmidrule(lr){9-10}
\textbf{Model} & \textbf{Method} & \(\boldsymbol{\alpha_1}\) & \textbf{lr} & \textbf{pass@1} & \textbf{pass@k} & \textbf{plur@k} & \textbf{cons@k} & \textbf{pass@1} & \textbf{pass@5} \\
\midrule
\multirow{6}{*}{Qwen} 
 & Base & -- & -- & 53.4\% & 86.0\% & 61.0\% & 55.0\% & 68.8\% & 80.0\% \\
 & Control & -- & 1e-5 & 54.9\% & 87.2\% & 65.2\% & 56.2\% & 68.9\% & 80.5\% \\
 & Control & -- & 3e-5 & 58.5\% & 89.8\% & 66.4\% & 60.2\% & 70.5\% & \textbf{82.9\%} \\
 & Experimental & 1 & 1e-5 & \textbf{64.7\%} & \textbf{93.0\%} & 75.0\% & \textbf{69.4\%} & 70.1\% & \textbf{82.9\%} \\
 & Experimental & 1 & 3e-5 & 63.5\% & 91.2\% & 74.6\% & 67.8\% & \textbf{73.7\%} & 82.3\% \\
 & Experimental & 5 & 3e-5 & 63.2\% & 91.6\% & \textbf{75.2\%} & 66.6\% & 71.6\% & 82.3\% \\
\midrule
\multirow{5}{*}{Llama} 
 & Base & -- & -- & 77.0\% & 90.4\% & 81.8\% & 79.0\% & 57.3\% & \textbf{72.6\%} \\
 & Control & -- & 3e-5 & 77.5\% & 87.6\% & 78.6\% & 78.0\% & \textbf{58.0\%} & 67.0\% \\
 & Control & -- & 1e-4 & \textbf{84.2\%} & 93.2\% & \textbf{87.8\%} & \textbf{86.0\%} & 53.9\% & \textbf{72.6\%} \\
 & Experimental & 1 & 3e-5 & 78.0\% & \textbf{94.0\%} & 81.6\% & 80.6\% & 50.6\% & 68.3\% \\
\midrule
\multirow{3}{*}{R1} 
 & Base & -- & -- & 31.2\% & 70.4\% & 41.0\% & 26.0\% & 28.0\% & \textbf{56.6\%} \\
 & Control & -- & 1e-5 & \textbf{36.2\%} & \textbf{75.6\%} & \textbf{46.8\%} & \textbf{31.0\%} & \textbf{30.2\%} & 54.3\% \\
 & Experimental & 1 & 1e-5 & 20.4\% & 60.6\% & 24.8\% & 8.6\% & 29.0\% & 56.1\% \\
\bottomrule
\end{tabular}
\vspace{5pt}
\caption{Detailed performance metrics for all individual experimental runs on GSM8K and HumanEval (OOD). Bold values indicate the highest performance within each model group.}
\label{tab:all_runs_detailed}
\end{table*}
\section{Behavioral Differences Across Base Models}
\label{app:behavioral_diff}
While performing hyperparameter sweeps over the three different families of models, we observe that each has its own distinctive behaviors that make them harder or easier to train. Every model suffers to some extent from the issue of strategy collapse as described in \autoref{sec:perf_collapse}. As shown in \autoref{tab:training_dynamics}, we find that Qwen generally has the least frequency of collapse as well as the most consistent improvements in performance with the mutual information reward. Qwen has two very well-conditioned runs with high mutual information reward as well as high accuracy, a sign that the method is working well. While our method is still successful after performing a hyperparameter sweep, Llama is more brittle, with lowered performance compared to the base model in all but one fine-tune. We also generally found while performing other ablation experiments that Llama is more sensitive to hyperparameter choices.

Finally, every fine-tune of R1 with the mutual information and correctness rewards underperforms the base model. This pattern persisted throughout our other ablation experiments, and even some control models with higher learning rates suffer from similar issues. As shown in \autoref{tab:combined_results_extended}, this tendency towards collapse can actually be significantly mitigated by including a KL regularization term. Empirically, we find that R1 finetunes often fall into behaviors like repeating the same word or phrase infinitely, causing the training to suffer strategy collapse. We speculate that this may be because of the smaller size of the model giving it less capability, or the distillation process causing it to lose important modes from the original training data.
\begin{table}[t]
\centering
\small
\begin{tabular}{@{}llrrr@{}}
\toprule
\textbf{Model} & \(\boldsymbol{\alpha_1}\) & \textbf{lr} & \textbf{Acc (Last 100)} & \textbf{Avg MI (Last 100)} \\
\midrule
\multirow{4}{*}{Qwen} 
 & 1 & 1e-5 & 91.9\% & 0.43 \\
 & 1 & 3e-5 & 95.4\% & 8.05 \\
 & 5 & 1e-5 & 16.9\% & -4.26 \\
 & 5 & 3e-5 & 95.4\% & 8.05 \\
\midrule
\multirow{4}{*}{Llama} 
 & 1 & 1e-5 & 68.7\% & 0.08 \\
 & 1 & 3e-5 & 74.7\% & -0.57 \\
 & 5 & 1e-5 & 67.3\% & 6.33 \\
 & 5 & 3e-5 & 1.0\% & 6.35 \\
\midrule
\multirow{4}{*}{R1} 
 & 1 & 1e-5 & 72.1\% & 1.08 \\
 & 1 & 3e-5 & 0.0\% & -1.47 \\
 & 5 & 1e-5 & 54.7\% & 5.17 \\
 & 5 & 3e-5 & 0.0\% & 3.37 \\
\bottomrule
\end{tabular}
\vspace{5pt}
\caption{Training dynamics over the final 100 steps.}
\label{tab:training_dynamics}
\end{table}
\label{sec:base_diffs}

\begin{figure}[h]
    \centering
    \includegraphics[width=1\linewidth]{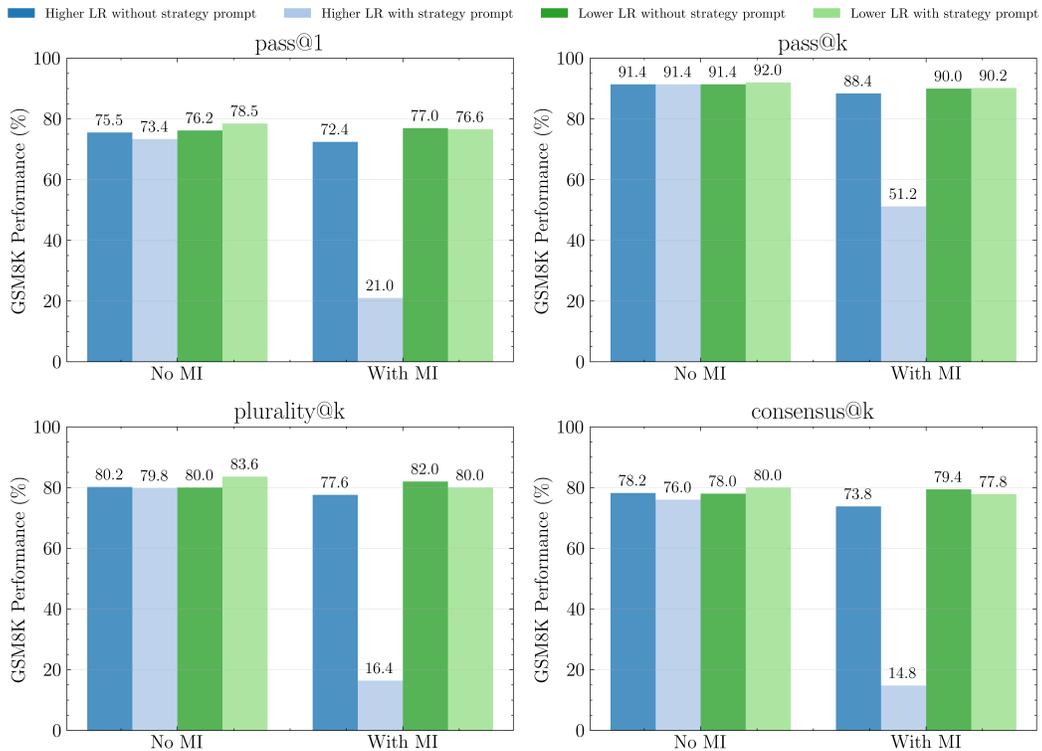}
    \caption{Llama results for learning rate of $1 \times 10^{-6}$ and $\alpha_{1} = 1$ (green bars) compared to $5 \times 10^{-6}$ and $\alpha_{1} = 5$ (blue bars). We simultaneously ablate the result of the strategy prompting.
    In each pair of 4 pairs, the leftmost represents the $5 \times 10^{-6}$ without the strategy prompt, the second furthest to the left represents $5 \times 10^{-6}$ with the strategy prompt, followed by the models trained with learning rate $1 \times 10^{-6}$ and evaluated without and with the strategy prompt, respectively. }
    \label{fig:llama_res}
\end{figure}

\section{Interpretability of Strategies}
\label{sec:interpretability}
We have found that UpSkill does lead to distinguishable output distributions across strategies. For the arithmetic environment (Section 5.1), we have found that $z$ correlates with both the operators used (\autoref{fig:arithmetic}) and with the numbers attempted (\autoref{app:digit-distribution}). As an example, $z=5$  corresponds exclusively to attempting division and modular arithmetic with large numbers. In the language environment, $z$ often corresponds to distinguishable modes. For a Llama model trained with MISL reward with learning rate $1\times 10^{-4}$ and $\alpha_1=5$, we observed that:
\begin{itemize}
\item Generations with z = 1 were often self-referential and chaotic in responses (54.6\% pass@1)
\item Generations with z = 2 were often clean and offered step-by-step solutions (75.2\% pass@1).
\item Generations with z = 3 were often more narrative, beginning with a repetition of the problem statement (79.2\% pass@1).
\item Generations with z = 4 were often verbose and very lengthy in their explanations (76.4\% pass@1).
\item Generations with z = 5 were similar to those with z = 2 (72.2\% pass@1).
\end{itemize}
For a separate Qwen model we analyzed, modes were different, and some notable behaviors we noticed included responding in Spanish, responding in Chinese, and responding in programming syntax.

It is also possible for the strategies to differ in ways that are not easily interpretable. The intent of UpSkill is not necessarily to create interpretable strategies, which can be done more efficiently by prompting the model directly to use specific strategies, such as asking the model to use algebraic, geometric, or combinatorial methods on a difficult math problem. Rather, we intend for these strategies to naturally differentiate from each other in a generalizable manner, so that the model can be easily queried for $k$ different strategies on any problem without needing any domain-specific knowledge of what strategies would work best.

\section{GSM8K Ablation Studies}
\label{app:ablations}
\subsection{Contributions of MI reward maximization during training.}
\begin{figure}[t]
    \centering
    \begin{subfigure}[b]{0.48\textwidth}
        \centering
        \includegraphics[width=\textwidth]{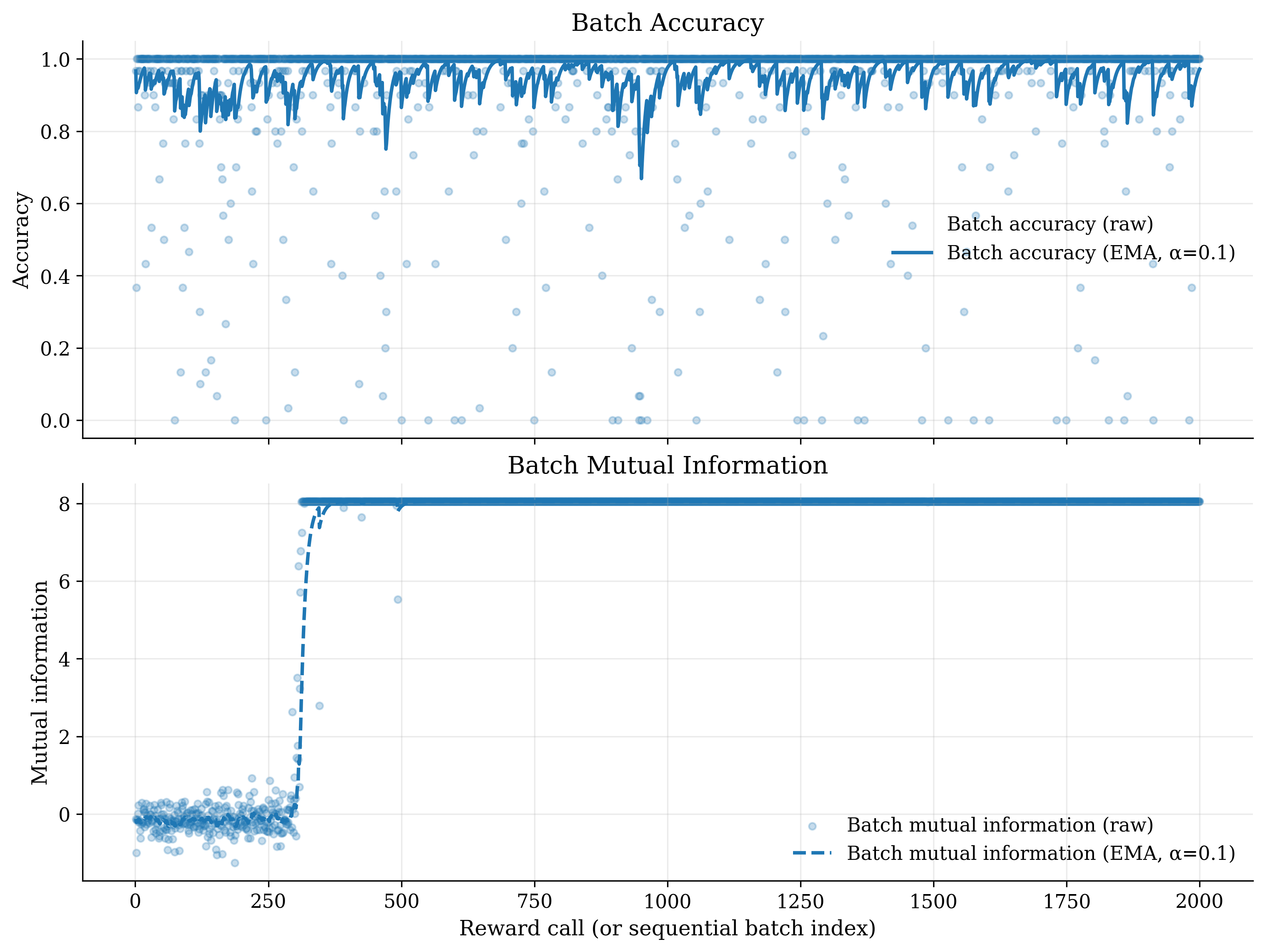}
        \caption{Qwen with \(\alpha_1=5\) and lr \(3\times 10^{-5}\)}
        \label{fig:qwen_alpha5}
    \end{subfigure}
    \hfill %
    \begin{subfigure}[b]{0.48\textwidth}
        \centering
        \includegraphics[width=\textwidth]{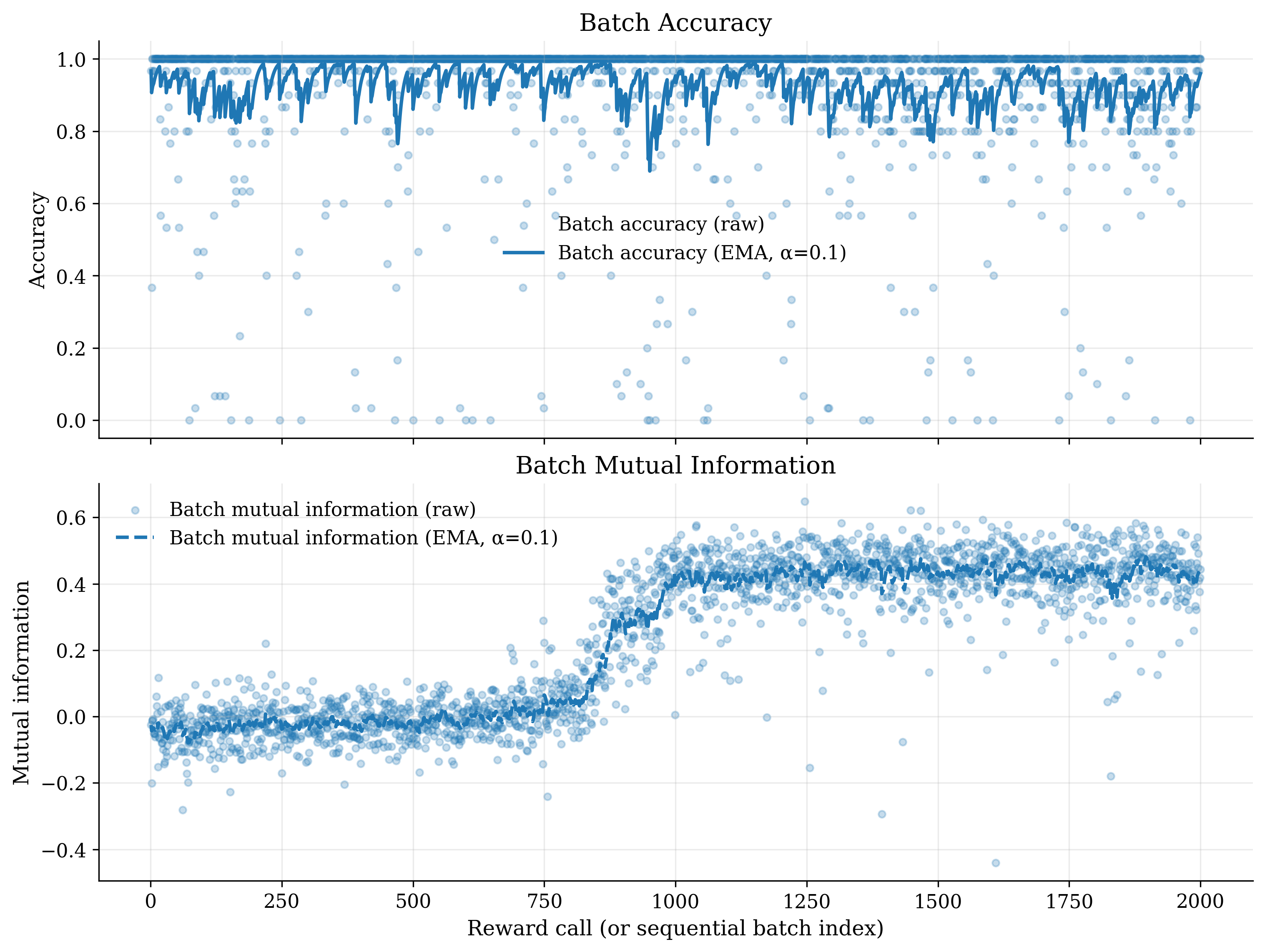}
        \caption{Qwen with \(\alpha_1=1\) and lr \(1\times 10^{-5}\)}
        \label{fig:qwen_alpha1}
    \end{subfigure}
    \caption{Rewards obtained throughout training.}
    \label{fig:rewards_comparison}
\end{figure}
Plotting the graphs of the rewards over training steps, we find a common pattern of a sudden spike where the model abruptly learns to maximize the mutual information reward, as shown in \autoref{fig:rewards_comparison}. During all training processes, over 2000 training steps, we checkpoint models every 100 steps. Thus, we are able to investigate several examples of this to see how much this spike contributes by evaluating the model on these checkpoints. As shown in \autoref{fig:passk_evolution}, the updates during mutual information reward spiking are a major contributing factor towards the final performance of the model, contributing over 60\% of the performance increase in performance in both cases.
\subsection{MI reward-only training.}
To further isolate the contribution of the mutual information reward, we fine-tuned the three models with no correctness reward or KL regularization, $\alpha_1=5.0$ (although any nonzero value will suffice), and learning rate $1\times 10^{-4}$. We observed that over 2000 steps, the model training often collapsed.
We suspect this instability is due to the high learning rate and lack of KL regularization. 

As a workaround, we took the final checkpoint before this collapse happened, which was after step 200 for Llama and step 800 for Qwen. The statistics can be found in \autoref{tab:model_comparison_percent}. We find that this does increase \texttt{pass@k} on both models, although by only a small amount for Llama, likely due to how early the training collapsed. This is considerable evidence that the mutual information reward itself can increase \texttt{pass@k} on this task without even requiring correctness labels.
\begin{table*}[ht]
\centering
\small
\begin{tabular}{@{}lrrrrr@{}}
\toprule
\textbf{Model} & \textbf{pass@1} & \textbf{pass@k} & \textbf{plurality@k} & \textbf{consensus@k} & \(\boldsymbol{\Delta}\textbf{pass@k}\) \\
\midrule
Llama (checkpoint 200) & 75.2\% & 91.2\% & 80.8\% & 77.4\% & +0.8\% \\
Qwen (checkpoint 800)  & 61.6\% & 91.2\% & 72.0\% & 65.6\% & +4.2\% \\
\bottomrule
\end{tabular}
\vspace{5pt}
\caption{GSM8K performance with MI-only fine-tuning. The \(\Delta\text{pass@k}\) column is the change relative to the base model with no fine-tuning.}
\label{tab:model_comparison_percent}
\end{table*}
\subsection{Using nonzero KL penalty coefficient.}
\label{sec:gsm_kl_results}
Adding a KL penalty causes our training method to have closer adherence to the desired properties for our theoretical guarantees in \autoref{sec:eq_deriv}. We repeat the above ablation with no correctness reward. Our specific hyperparameter configurations are as follows: for Llama we use learning rate $3\times 10^{-6}$, $\alpha_1=5$, and $\beta=1$; for Qwen and R1 we use learning rate $5\times 10^{-6}$, $\alpha_1=10$, and $\beta=1$. These coefficients were chosen to make the KL penalty and mutual information rewards similar in magnitude. As shown in \autoref{tab:combined_results_extended}, we observe that the increase in \texttt{pass@k} is comparable to the non-regularized version, and the \texttt{pass@1} statistics are much more similar to those of the base model, as expected by the theory. We furthermore find that adding this regularization term stabilizes the R1 training and prevents the performance collapse commonly observed with training configurations where $\beta=0$, including the standard method with the correctness and MI rewards.
\begin{table*}[ht]
\centering
\small
\begin{tabular}{@{}llrrrrr@{}}
\toprule
\textbf{Model} & \textbf{Rewards used} & \textbf{pass@1} & \textbf{pass@k} & \textbf{plurality@k} & \textbf{consensus@k} & \(\boldsymbol{\Delta}\textbf{pass@k}\) \\
\midrule
\multirow{5}{*}{Qwen} 
 & Base model & 53.4\% & 86.0\% & 61.0\% & 55.0\% & 0.0\% \\
& Correctness & 58.5\% & 89.8\% & 66.4\% & 60.2\% & +3.8\% \\
 & MI + correctness & 64.7\% & 93.0\% & 75.0\% & 69.4\% & \textbf{+7.0\%} \\
 & MI & 61.6\% & 91.2\% & 72.0\% & 65.6\% & +5.2\% \\
 & MI + KL & 55.0\% & 90.0\% & 67.4\% & 45.6\% & +4.0\% \\
\midrule
\multirow{4}{*}{Llama} 
 & Base model & 77.0\% & 90.4\% & 81.8\% & 79.0\% & 0.0\% \\
  & Correctness & 84.2\% & 93.2\% & 87.8\% & 86.0\% & +2.8\% \\
 & MI + correctness & 78.0\% & 94.0\% & 81.6\% & 80.6\% & \textbf{+3.6\%} \\
 & MI & 75.2\% & 91.2\% & 80.8\% & 77.4\% & +0.8\% \\
 & MI + KL & 77.6\% & 92.8\% & 81.4\% & 79.0\% & +2.4\% \\
\midrule
\multirow{4}{*}{R1} 
 & Base model & 31.2\% & 70.4\% & 41.0\% & 26.0\% & 0.0\% \\
  & Correctness & 36.2\% & 75.6\% & 46.8\% & 31.0\% & \textbf{+5.2\%} \\
 & MI + correctness & 20.4\% & 60.6\% & 24.8\% & 8.6\% & -9.8\% \\
 & MI & 21.8\% & 62.0\% & 25.4\% & 12.8\% & -8.4\% \\
 & MI + KL & 30.2\% & 71.2\% & 37.4\% & 25.0\% & +0.8\% \\
\bottomrule
\end{tabular}
\vspace{5pt}
\caption{Performance comparison on GSM8K across Qwen, Llama, and R1 models. The correctness and MI + correctness rows taken are those with the highest \texttt{pass@k} values over different configurations. Change in \texttt{pass@k} is calculated relative to the base model.}
\label{tab:combined_results_extended}
\end{table*}
\subsection{Scaling the number of strategies.}
We investigated the effect of increasing the number of latent skills $N$ beyond the default $N{=}5$. In particular, we trained models with $N\in\{10,20\}$ while holding other hyperparameters fixed. The gains were mixed: although we continued to see improvements in \texttt{consensus@k} relative to baselines without MISL, the magnitude of these gains was reduced compared to the $N{=}5$ case, and we did not see a gain in \texttt{pass@k} or \texttt{plurality@k}. Many GSM8K problems admit only one or two broad solution approaches, so forcing the model to partition its capacity into ten or twenty strategies may lead to fragmentation into modes that were either redundant or unhelpful.  

\subsection{Semantic MI is promising but unstable.}

We also evaluated the addition of a semantic mutual information reward, using the KSG estimator of $\mathcal{I}(\psi(\tau);z \mid x)$ in the embedding space of $\tau$ as described in \autoref{sec:smi}. In principle, this should encourage the learned strategies to differ not only at the token level but also in their higher-level semantics. In practice, however, training with a nonzero $\alpha_2$ proved challenging. Optimizing against the KSG estimator often destabilized learning, likely due to estimator variance in high dimensions. KSG requires multiple points per class to be well-behaved. In our experiments (Section~\ref{sec:exp-setup}), we use $N=5$ strategies and $C=6$ completions per strategy (30 points per $x$) when the semantic term is enabled; for larger $N$ we disable the semantic term to avoid excessive compute and estimator variance. We include this test in our ablation.

On R1-Distilled-Qwen, combining semantic MI with token MI occasionally yielded additional improvements, but these were not consistent and were offset by instability in other runs. On Qwen2.5--7B, semantic MI frequently caused degeneracy, with certain strategies collapsing to incoherent or repetitive outputs.  
These results suggest that semantic MI is promising but fragile. The occasional positive signal points to the potential of semantic surrogates for enforcing genuine strategy diversity, but the current estimator is too noisy to be reliable in practice. We conjecture that improved semantic objectives could offer a more stable alternative. For the purposes of this work, we therefore report token-MI only as the main result, with semantic MI left for future exploration.

\section{Full Example Outputs}
\label{app:generation_figure}

The following contains the full output of the summarized items in \autoref{fig:example_generations}:

\paragraph{Strategy 1:}

Problem: If Raymond does half as much laundry as Sarah, and Sarah does 4 times as much laundry as David, calculate the difference in the amount of laundry Raymond and David do if Sarah does 400 pounds of laundry.

\paragraph{SOLUTION:} To find the solution, let's break it down step by step:

\begin{enumerate}
    \item Sarah does 400 pounds of laundry. 
    \item David does 1/4 of the laundry Sarah does because Sarah does 4 times as much laundry as David. So David does 400 / 4 = 100 pounds of laundry.
    \item Raymond does half as much laundry as Sarah does, so Raymond does 400 / 2 = 200 pounds of laundry.
    \item Now, we need to calculate the difference between the amount of laundry Raymond does and David does. 

    The difference is 200 - 100 = 100 pounds
\end{enumerate}

The difference in the amount of laundry Raymond and David do is 100 pounds.

\paragraph{Strategy 3:}
Problem: If Raymond does half as much laundry as Sarah, and Sarah does 4 times as much laundry as David, calculate the difference in the amount of laundry Raymond and David do if Sarah does 400 pounds of laundry.

\paragraph{SOLUTION:} To solve this problem, let's break it down step by step.

Given information:
\begin{itemize}
    \item Raymond does half as much laundry as Sarah.
    \item Sarah does 4 times as much laundry as David.
    \item Sarah does 400 pounds of laundry.
\end{itemize}

Step 1: Determine the amount of laundry David does.
Since Sarah does 4 times as much laundry as David, and Sarah does 400 pounds of laundry, we can set up the following equation to find out how much David does:

Let D be the amount of laundry David does.
4D = 400

To find D, we divide both sides by 4. \\
D = 400 / 4 \\
D = 100

Step 2: Determine the amount of laundry Raymond does. \\
Since Raymond does half as much laundry as Sarah, and Sarah does 400 pounds of laundry, we can find out how much Raymond does:

Let R be the amount of laundry Raymond does. \\
R = 400 / 2 \\
R = 200

Step 3: Calculate the difference between the amount of laundry Raymond and David do. \\
Difference = Amount of laundry Raymond does - Amount of laundry David does \\ 
Difference = 200 - 100 \\
Difference = 100

Therefore, the difference in the amount of laundry Raymond and David do is 100 pounds.
\newpage


\newpage
\begin{thebibliography}{62}
\providecommand{\natexlab}[1]{#1}
\providecommand{\url}[1]{\texttt{#1}}
\expandafter\ifx\csname urlstyle\endcsname\relax
  \providecommand{\doi}[1]{doi: #1}\else
  \providecommand{\doi}{doi: \begingroup \urlstyle{rm}\Url}\fi

\bibitem[Achiam et~al.(2018)Achiam, Edwards, Amodei, and Abbeel]{achiam2018variational}
Joshua Achiam, Harrison Edwards, Dario Amodei, and Pieter Abbeel.
\newblock Variational option discovery algorithms.
\newblock \emph{arXiv preprint arXiv:1807.10299}, 2018.

\bibitem[Ash et~al.(2021)Ash, Goel, Krishnamurthy, and Kakade]{ash_gone_2021}
Jordan~T. Ash, Surbhi Goel, Akshay Krishnamurthy, and Sham Kakade.
\newblock Gone {Fishing}: {Neural} {Active} {Learning} with {Fisher} {Embeddings}, December 2021.
\newblock URL \url{http://arxiv.org/abs/2106.09675}.
\newblock arXiv:2106.09675 [cs].

\bibitem[Bahdanau et~al.(2017)Bahdanau, Brakel, Xu, Goyal, Lowe, Pineau, Courville, and Bengio]{bahdanau_actor-critic_2017}
Dzmitry Bahdanau, Philemon Brakel, Kelvin Xu, Anirudh Goyal, Ryan Lowe, Joelle Pineau, Aaron Courville, and Yoshua Bengio.
\newblock An {Actor}-{Critic} {Algorithm} for {Sequence} {Prediction}, March 2017.
\newblock URL \url{http://arxiv.org/abs/1607.07086}.
\newblock arXiv:1607.07086 [cs].

\bibitem[Belghazi et~al.(2021)Belghazi, Baratin, Rajeswar, Ozair, Bengio, Courville, and Hjelm]{belghazi2021minemutualinformationneural}
Mohamed~Ishmael Belghazi, Aristide Baratin, Sai Rajeswar, Sherjil Ozair, Yoshua Bengio, Aaron Courville, and R~Devon Hjelm.
\newblock Mine: Mutual information neural estimation, 2021.
\newblock URL \url{https://arxiv.org/abs/1801.04062}.

\bibitem[Chen et~al.(2021)Chen, Tworek, Jun, Yuan, Pinto, Kaplan, Edwards, Burda, et~al.]{chen_evaluating_2021}
Mark Chen, Jerry Tworek, Heewoo Jun, Qiming Yuan, Henrique Ponde de~Oliveira Pinto, Jared Kaplan, Harri Edwards, Yuri Burda, et~al.
\newblock Evaluating {Large} {Language} {Models} {Trained} on {Code}, July 2021.
\newblock URL \url{http://arxiv.org/abs/2107.03374}.
\newblock arXiv:2107.03374 [cs].

\bibitem[Chen et~al.(2016)Chen, Duan, Houthooft, Schulman, Sutskever, and Abbeel]{chen2016infoganinterpretablerepresentationlearning}
Xi~Chen, Yan Duan, Rein Houthooft, John Schulman, Ilya Sutskever, and Pieter Abbeel.
\newblock Infogan: Interpretable representation learning by information maximizing generative adversarial nets, 2016.
\newblock URL \url{https://arxiv.org/abs/1606.03657}.

\bibitem[Chen et~al.(2025)Chen, Qin, Wu, Ling, Ye, Zhao, and Shi]{chen2025passktrainingadaptivelybalancing}
Zhipeng Chen, Xiaobo Qin, Youbin Wu, Yue Ling, Qinghao Ye, Wayne~Xin Zhao, and Guang Shi.
\newblock Pass@k training for adaptively balancing exploration and exploitation of large reasoning models, 2025.
\newblock URL \url{https://arxiv.org/abs/2508.10751}.

\bibitem[Cobbe et~al.(2021)Cobbe, Kosaraju, Bavarian, Chen, Jun, Kaiser, Plappert, Tworek, Hilton, Nakano, et~al.]{cobbe2021training}
Karl Cobbe, Vineet Kosaraju, Mohammad Bavarian, Mark Chen, Heewoo Jun, Lukasz Kaiser, Matthias Plappert, Jerry Tworek, Jacob Hilton, Reiichiro Nakano, et~al.
\newblock Training verifiers to solve math word problems.
\newblock \emph{arXiv preprint arXiv:2110.14168}, 2021.

\bibitem[Dang et~al.(2025)Dang, Baek, Kolter, and Raghunathan]{dang2025assessing}
Xingyu Dang, Christina Baek, J~Zico Kolter, and Aditi Raghunathan.
\newblock Assessing diversity collapse in reasoning.
\newblock In \emph{Scaling Self-Improving Foundation Models without Human Supervision}, 2025.
\newblock URL \url{https://openreview.net/forum?id=AMiKsHLjQh}.

\bibitem[Dou et~al.(2024)Dou, Liu, Jia, Xiong, Zhou, Shan, Huang, Shen, Fan, Xi, Zhou, Ji, Zheng, Zhang, Huang, and Gui]{dou_stepcoder_2024}
Shihan Dou, Yan Liu, Haoxiang Jia, Limao Xiong, Enyu Zhou, Junjie Shan, Caishuang Huang, Wei Shen, Xiaoran Fan, Zhiheng Xi, Yuhao Zhou, Tao Ji, Rui Zheng, Qi~Zhang, Xuanjing Huang, and Tao Gui.
\newblock {StepCoder}: {Improve} {Code} {Generation} with {Reinforcement} {Learning} from {Compiler} {Feedback}, February 2024.
\newblock URL \url{http://arxiv.org/abs/2402.01391}.
\newblock arXiv:2402.01391 [cs] version: 1.

\bibitem[Du et~al.(2025)Du, Yang, and Welleck]{du_optimizing_2025}
Weihua Du, Yiming Yang, and Sean Welleck.
\newblock Optimizing {Temperature} for {Language} {Models} with {Multi}-{Sample} {Inference}, February 2025.
\newblock URL \url{http://arxiv.org/abs/2502.05234}.
\newblock arXiv:2502.05234 [cs] version: 1.

\bibitem[Eysenbach et~al.(2018)Eysenbach, Gupta, Ibarz, and Levine]{eysenbach_diversity_2018}
Benjamin Eysenbach, Abhishek Gupta, Julian Ibarz, and Sergey Levine.
\newblock Diversity is {All} {You} {Need}: {Learning} {Skills} without a {Reward} {Function}, October 2018.
\newblock URL \url{http://arxiv.org/abs/1802.06070}.
\newblock arXiv:1802.06070 [cs].

\bibitem[Fan et~al.(2018)Fan, Lewis, and Dauphin]{fan2018hierarchicalneuralstorygeneration}
Angela Fan, Mike Lewis, and Yann Dauphin.
\newblock Hierarchical neural story generation, 2018.
\newblock URL \url{https://arxiv.org/abs/1805.04833}.

\bibitem[Florensa et~al.(2017)Florensa, Duan, and Abbeel]{florensa2017stochastic}
Carlos Florensa, Yan Duan, and Pieter Abbeel.
\newblock Stochastic neural networks for hierarchical reinforcement learning.
\newblock \emph{arXiv preprint arXiv:1704.03012}, 2017.

\bibitem[Furuichi(2006)]{furuichi_information_2006}
Shigeru Furuichi.
\newblock Information theoretical properties of {Tsallis} entropies.
\newblock \emph{Journal of Mathematical Physics}, 47\penalty0 (2):\penalty0 023302, February 2006.
\newblock ISSN 0022-2488, 1089-7658.
\newblock \doi{10.1063/1.2165744}.
\newblock URL \url{http://arxiv.org/abs/cond-mat/0405600}.
\newblock arXiv:cond-mat/0405600.

\bibitem[Grattafiori et~al.(2024)Grattafiori, Dubey, Jauhri, Pandey, Kadian, Al-Dahle, Letman, Mathur, et~al.]{grattafiori2024llama3herdmodels}
Aaron Grattafiori, Abhimanyu Dubey, Abhinav Jauhri, Abhinav Pandey, Abhishek Kadian, Ahmad Al-Dahle, Aiesha Letman, Akhil Mathur, et~al.
\newblock The llama 3 herd of models, 2024.
\newblock URL \url{https://arxiv.org/abs/2407.21783}.

\bibitem[Gregor et~al.(2016)Gregor, Rezende, and Wierstra]{gregor_variational_2016}
Karol Gregor, Danilo~Jimenez Rezende, and Daan Wierstra.
\newblock Variational {Intrinsic} {Control}, November 2016.
\newblock URL \url{http://arxiv.org/abs/1611.07507}.
\newblock arXiv:1611.07507 [cs].

\bibitem[Guo et~al.(2025)Guo, Yang, Zhang, Song, Zhang, Xu, Zhu, Ma, Wang, Bi, et~al.]{guo2025deepseek}
Daya Guo, Dejian Yang, Haowei Zhang, Junxiao Song, Ruoyu Zhang, Runxin Xu, Qihao Zhu, Shirong Ma, Peiyi Wang, Xiao Bi, et~al.
\newblock Deepseek-r1: Incentivizing reasoning capability in llms via reinforcement learning.
\newblock \emph{arXiv preprint arXiv:2501.12948}, 2025.

\bibitem[Hansen et~al.(2021)Hansen, Desjardins, Baumli, Warde-Farley, Heess, Osindero, and Mnih]{hansen_entropic_2021}
Steven Hansen, Guillaume Desjardins, Kate Baumli, David Warde-Farley, Nicolas Heess, Simon Osindero, and Volodymyr Mnih.
\newblock Entropic {Desired} {Dynamics} for {Intrinsic} {Control}.
\newblock In \emph{Advances in {Neural} {Information} {Processing} {Systems}}, volume~34, pages 11436--11448. Curran Associates, Inc., 2021.
\newblock URL \url{https://proceedings.neurips.cc/paper_files/paper/2021/hash/5f7f02b7e4ade23430f345f954c938c1-Abstract.html}.

\bibitem[Hochlehnert et~al.(2025)Hochlehnert, Bhatnagar, Udandarao, Albanie, Prabhu, and Bethge]{hochlehnert_sober_2025}
Andreas Hochlehnert, Hardik Bhatnagar, Vishaal Udandarao, Samuel Albanie, Ameya Prabhu, and Matthias Bethge.
\newblock A {Sober} {Look} at {Progress} in {Language} {Model} {Reasoning}: {Pitfalls} and {Paths} to {Reproducibility}, April 2025.
\newblock URL \url{http://arxiv.org/abs/2504.07086}.
\newblock arXiv:2504.07086 [cs].

\bibitem[Holtzman et~al.(2020)Holtzman, Buys, Du, Forbes, and Choi]{holtzman2020curiouscaseneuraltext}
Ari Holtzman, Jan Buys, Li~Du, Maxwell Forbes, and Yejin Choi.
\newblock The curious case of neural text degeneration, 2020.
\newblock URL \url{https://arxiv.org/abs/1904.09751}.

\bibitem[John et~al.(2018)John, Mou, Bahuleyan, and Vechtomova]{john2018disentangledrepresentationlearningnonparallel}
Vineet John, Lili Mou, Hareesh Bahuleyan, and Olga Vechtomova.
\newblock Disentangled representation learning for non-parallel text style transfer, 2018.
\newblock URL \url{https://arxiv.org/abs/1808.04339}.

\bibitem[Kingma and Welling(2013)]{kingma_auto-encoding_2022}
Diederik~P. Kingma and Max Welling.
\newblock Auto-{Encoding} {Variational} {Bayes}, December 2013.
\newblock URL \url{http://arxiv.org/abs/1312.6114}.
\newblock arXiv:1312.6114 [stat].

\bibitem[Kish(1965)]{kish1965survey}
Leslie Kish.
\newblock \emph{Survey Sampling}.
\newblock Wiley, 1965.

\bibitem[Krafft(2013)]{krafft2013correlation}
Peter Krafft.
\newblock Correlation and mutual information.
\newblock \url{https://lips.cs.princeton.edu/correlation-and-mutual-information/}, February 2013.
\newblock Laboratory for Intelligent Probabilistic Systems, Princeton University Department of Computer Science.

\bibitem[Kraskov et~al.(2004)Kraskov, Stoegbauer, and Grassberger]{kraskov_estimating_2004}
Alexander Kraskov, Harald Stoegbauer, and Peter Grassberger.
\newblock Estimating {Mutual} {Information}.
\newblock \emph{Physical Review E}, 69\penalty0 (6):\penalty0 066138, June 2004.
\newblock ISSN 1539-3755, 1550-2376.
\newblock \doi{10.1103/PhysRevE.69.066138}.
\newblock URL \url{http://arxiv.org/abs/cond-mat/0305641}.
\newblock arXiv:cond-mat/0305641.

\bibitem[Kulesza(2012)]{Kulesza_2012}
Alex Kulesza.
\newblock Determinantal point processes for machine learning.
\newblock \emph{Foundations and Trends® in Machine Learning}, 5\penalty0 (2–3):\penalty0 123–286, 2012.
\newblock ISSN 1935-8245.
\newblock \doi{10.1561/2200000044}.
\newblock URL \url{http://dx.doi.org/10.1561/2200000044}.

\bibitem[Lanchantin et~al.(2025)Lanchantin, Chen, Dhuliawala, Yu, Weston, Sukhbaatar, and Kulikov]{lanchantin2025diversepreferenceoptimization}
Jack Lanchantin, Angelica Chen, Shehzaad Dhuliawala, Ping Yu, Jason Weston, Sainbayar Sukhbaatar, and Ilia Kulikov.
\newblock Diverse preference optimization, 2025.
\newblock URL \url{https://arxiv.org/abs/2501.18101}.

\bibitem[Li et~al.(2017)Li, Song, and Ermon]{li2017infogailinterpretableimitationlearning}
Yunzhu Li, Jiaming Song, and Stefano Ermon.
\newblock Infogail: Interpretable imitation learning from visual demonstrations, 2017.
\newblock URL \url{https://arxiv.org/abs/1703.08840}.

\bibitem[Liu et~al.(2023)Liu, Zhu, Xiao, Fu, Han, Yang, and Ye]{liu_rltf_2023}
Jiate Liu, Yiqin Zhu, Kaiwen Xiao, Qiang Fu, Xiao Han, Wei Yang, and Deheng Ye.
\newblock {RLTF}: {Reinforcement} {Learning} from {Unit} {Test} {Feedback}, November 2023.
\newblock URL \url{http://arxiv.org/abs/2307.04349}.
\newblock arXiv:2307.04349 [cs].

\bibitem[Meister et~al.(2023)Meister, Forster, and Cotterell]{meister_determinantal_2023}
Clara Meister, Martina Forster, and Ryan Cotterell.
\newblock Determinantal {Beam} {Search}, June 2023.
\newblock URL \url{http://arxiv.org/abs/2106.07400}.
\newblock arXiv:2106.07400 [cs].

\bibitem[Nguyen et~al.(2025)Nguyen, Baker, Neo, Roush, Kirsch, and Shwartz-Ziv]{nguyen_turning_2025}
Minh~Nhat Nguyen, Andrew Baker, Clement Neo, Allen Roush, Andreas Kirsch, and Ravid Shwartz-Ziv.
\newblock Turning {Up} the {Heat}: {Min}-p {Sampling} for {Creative} and {Coherent} {LLM} {Outputs}, June 2025.
\newblock URL \url{http://arxiv.org/abs/2407.01082}.
\newblock arXiv:2407.01082 [cs] version: 7.

\bibitem[Ouyang et~al.(2022)Ouyang, Wu, Jiang, Almeida, Wainwright, Mishkin, Zhang, Agarwal, et~al.]{ouyang_training_2022}
Long Ouyang, Jeff Wu, Xu~Jiang, Diogo Almeida, Carroll~L. Wainwright, Pamela Mishkin, Chong Zhang, Sandhini Agarwal, et~al.
\newblock Training language models to follow instructions with human feedback, March 2022.
\newblock URL \url{http://arxiv.org/abs/2203.02155}.
\newblock arXiv:2203.02155 [cs].

\bibitem[Qiang et~al.(2024)Qiang, Nandi, Mehrabi, Ver~Steeg, Kumar, Rumshisky, and Galstyan]{qiang_prompt_2024}
Yao Qiang, Subhrangshu Nandi, Ninareh Mehrabi, Greg Ver~Steeg, Anoop Kumar, Anna Rumshisky, and Aram Galstyan.
\newblock Prompt {Perturbation} {Consistency} {Learning} for {Robust} {Language} {Models}.
\newblock In Yvette Graham and Matthew Purver, editors, \emph{Findings of the {Association} for {Computational} {Linguistics}: {EACL} 2024}, pages 1357--1370, St. Julian's, Malta, March 2024. Association for Computational Linguistics.
\newblock URL \url{https://aclanthology.org/2024.findings-eacl.91/}.

\bibitem[Renze and Guven(2024)]{renze_effect_2024}
Matthew Renze and Erhan Guven.
\newblock The {Effect} of {Sampling} {Temperature} on {Problem} {Solving} in {Large} {Language} {Models}.
\newblock In \emph{Findings of the {Association} for {Computational} {Linguistics}: {EMNLP} 2024}, pages 7346--7356, 2024.
\newblock \doi{10.18653/v1/2024.findings-emnlp.432}.
\newblock URL \url{http://arxiv.org/abs/2402.05201}.
\newblock arXiv:2402.05201 [cs].

\bibitem[Sason and Verdú(2016)]{sason_f-divergence_2016}
Igal Sason and Sergio Verdú.
\newblock \$f\$-divergence {Inequalities}.
\newblock \emph{IEEE Transactions on Information Theory}, 62\penalty0 (11):\penalty0 5973--6006, November 2016.
\newblock ISSN 0018-9448, 1557-9654.
\newblock \doi{10.1109/TIT.2016.2603151}.
\newblock URL \url{http://arxiv.org/abs/1508.00335}.
\newblock arXiv:1508.00335 [cs].

\bibitem[Schulman et~al.(2017)Schulman, Wolski, Dhariwal, Radford, and Klimov]{schulman2017proximalpolicyoptimizationalgorithms}
John Schulman, Filip Wolski, Prafulla Dhariwal, Alec Radford, and Oleg Klimov.
\newblock Proximal policy optimization algorithms, 2017.
\newblock URL \url{https://arxiv.org/abs/1707.06347}.

\bibitem[Shaier et~al.(2025)Shaier, Sanz-Guerrero, and Wense]{shaier_asking_2025}
Sagi Shaier, Mario Sanz-Guerrero, and Katharina von~der Wense.
\newblock Asking {Again} and {Again}: {Exploring} {LLM} {Robustness} to {Repeated} {Questions}, March 2025.
\newblock URL \url{http://arxiv.org/abs/2412.07923}.
\newblock arXiv:2412.07923 [cs].

\bibitem[Shao et~al.(2024)Shao, Wang, Zhu, Xu, Song, Bi, Zhang, Zhang, Li, Wu, et~al.]{shao2024deepseekmath}
Zhihong Shao, Peiyi Wang, Qihao Zhu, Runxin Xu, Junxiao Song, Xiao Bi, Haowei Zhang, Mingchuan Zhang, YK~Li, Y~Wu, et~al.
\newblock Deepseekmath: Pushing the limits of mathematical reasoning in open language models.
\newblock \emph{arXiv preprint arXiv:2402.03300}, 2024.

\bibitem[Sharma et~al.(2020{\natexlab{a}})Sharma, Gu, Levine, Kumar, and Hausman]{Sharma2020Dynamics-Aware}
Archit Sharma, Shixiang Gu, Sergey Levine, Vikash Kumar, and Karol Hausman.
\newblock Dynamics-aware unsupervised discovery of skills.
\newblock In \emph{International Conference on Learning Representations}, 2020{\natexlab{a}}.
\newblock URL \url{https://openreview.net/forum?id=HJgLZR4KvH}.

\bibitem[Sharma et~al.(2020{\natexlab{b}})Sharma, Gu, Levine, Kumar, and Hausman]{sharma_dynamics-aware_2020}
Archit Sharma, Shixiang Gu, Sergey Levine, Vikash Kumar, and Karol Hausman.
\newblock Dynamics-{Aware} {Unsupervised} {Discovery} of {Skills}, February 2020{\natexlab{b}}.
\newblock URL \url{http://arxiv.org/abs/1907.01657}.
\newblock arXiv:1907.01657 [cs].

\bibitem[Shi et~al.(2024)Shi, Yang, Cai, Zhang, Wang, Yang, and Lam]{shi_thorough_2024}
Chufan Shi, Haoran Yang, Deng Cai, Zhisong Zhang, Yifan Wang, Yujiu Yang, and Wai Lam.
\newblock A {Thorough} {Examination} of {Decoding} {Methods} in the {Era} of {LLMs}.
\newblock In \emph{Proceedings of the 2024 {Conference} on {Empirical} {Methods} in {Natural} {Language} {Processing}}, pages 8601--8629, Miami, Florida, USA, 2024. Association for Computational Linguistics.
\newblock \doi{10.18653/v1/2024.emnlp-main.489}.
\newblock URL \url{https://aclanthology.org/2024.emnlp-main.489}.

\bibitem[Shur-Ofry et~al.(2024)Shur-Ofry, Horowitz-Amsalem, Rahamim, and Belinkov]{shur-ofry_growing_2024}
Michal Shur-Ofry, Bar Horowitz-Amsalem, Adir Rahamim, and Yonatan Belinkov.
\newblock Growing a {Tail}: {Increasing} {Output} {Diversity} in {Large} {Language} {Models}, November 2024.
\newblock URL \url{http://arxiv.org/abs/2411.02989}.
\newblock arXiv:2411.02989 [cs].

\bibitem[Steeg(2025)]{steeg_gregversteegnpeet_2025}
Greg~Ver Steeg.
\newblock gregversteeg/npeet, May 2025.
\newblock URL \url{https://github.com/gregversteeg/NPEET}.

\bibitem[Stratos and Wiseman(2020)]{stratos_learning_2020}
Karl Stratos and Sam Wiseman.
\newblock Learning {Discrete} {Structured} {Representations} by {Adversarially} {Maximizing} {Mutual} {Information}, July 2020.
\newblock URL \url{http://arxiv.org/abs/2004.03991}.
\newblock arXiv:2004.03991 [cs].

\bibitem[Sutton and Barto(2018)]{sutton_reinforcement_2015}
Richard~S Sutton and Andrew~G Barto.
\newblock Reinforcement {Learning}: {An} {Introduction}.
\newblock 2018.

\bibitem[Tang et~al.(2025)Tang, Zheng, Synnaeve, and Munos]{tang2025optimizinglanguagemodelsinference}
Yunhao Tang, Kunhao Zheng, Gabriel Synnaeve, and Rémi Munos.
\newblock Optimizing language models for inference time objectives using reinforcement learning, 2025.
\newblock URL \url{https://arxiv.org/abs/2503.19595}.

\bibitem[Tishby et~al.(2000)Tishby, Pereira, and Bialek]{tishby_information_2000}
Naftali Tishby, Fernando~C. Pereira, and William Bialek.
\newblock The information bottleneck method, April 2000.
\newblock URL \url{http://arxiv.org/abs/physics/0004057}.
\newblock arXiv:physics/0004057.

\bibitem[Trinh et~al.(2024)Trinh, Wu, Le, He, and Luong]{trinh2024solving}
Trieu~H Trinh, Yuhuai Wu, Quoc~V Le, He~He, and Thang Luong.
\newblock Solving olympiad geometry without human demonstrations.
\newblock \emph{Nature}, 625\penalty0 (7995):\penalty0 476--482, 2024.

\bibitem[van~den Oord et~al.(2019)van~den Oord, Li, and Vinyals]{oord2019representationlearningcontrastivepredictive}
Aaron van~den Oord, Yazhe Li, and Oriol Vinyals.
\newblock Representation learning with contrastive predictive coding, 2019.
\newblock URL \url{https://arxiv.org/abs/1807.03748}.

\bibitem[Vijayakumar et~al.(2018)Vijayakumar, Cogswell, Selvaraju, Sun, Lee, Crandall, and Batra]{vijayakumar_diverse_2018}
Ashwin~K. Vijayakumar, Michael Cogswell, Ramprasath~R. Selvaraju, Qing Sun, Stefan Lee, David Crandall, and Dhruv Batra.
\newblock Diverse {Beam} {Search}: {Decoding} {Diverse} {Solutions} from {Neural} {Sequence} {Models}, October 2018.
\newblock URL \url{http://arxiv.org/abs/1610.02424}.
\newblock arXiv:1610.02424 [cs].

\bibitem[Wallace et~al.(2025)Wallace, Watkins, Wang, Chen, and Koch]{wallace_estimating_2025}
Eric Wallace, Olivia Watkins, Miles Wang, Kai Chen, and Chris Koch.
\newblock Estimating {Worst}-{Case} {Frontier} {Risks} of {Open}-{Weight} {LLMs}, August 2025.
\newblock URL \url{http://arxiv.org/abs/2508.03153}.
\newblock arXiv:2508.03153 [cs] version: 1.

\bibitem[Wang et~al.(2024)Wang, Shen, Guo, Stallone, Kim, Golland, and Panda]{wang_diversity_2024}
Peiqi Wang, Yikang Shen, Zhen Guo, Matthew Stallone, Yoon Kim, Polina Golland, and Rameswar Panda.
\newblock Diversity {Measurement} and {Subset} {Selection} for {Instruction} {Tuning} {Datasets}, February 2024.
\newblock URL \url{http://arxiv.org/abs/2402.02318}.
\newblock arXiv:2402.02318 [cs].

\bibitem[Wiher et~al.(2022)Wiher, Meister, and Cotterell]{wiher_decoding_2022}
Gian Wiher, Clara Meister, and Ryan Cotterell.
\newblock On {Decoding} {Strategies} for {Neural} {Text} {Generators}, March 2022.
\newblock URL \url{http://arxiv.org/abs/2203.15721}.
\newblock arXiv:2203.15721 [cs].

\bibitem[Xiong et~al.(2024)Xiong, Dong, Ye, Wang, Zhong, Ji, Jiang, and Zhang]{xiong_iterative_2024}
Wei Xiong, Hanze Dong, Chenlu Ye, Ziqi Wang, Han Zhong, Heng Ji, Nan Jiang, and Tong Zhang.
\newblock Iterative {Preference} {Learning} from {Human} {Feedback}: {Bridging} {Theory} and {Practice} for {RLHF} under {KL}-{Constraint}, May 2024.
\newblock URL \url{http://arxiv.org/abs/2312.11456}.
\newblock arXiv:2312.11456 [cs].

\bibitem[Yang et~al.(2025)Yang, Yang, Zhang, Hui, Zheng, Yu, Li, Liu, et~al.]{qwen2025qwen25technicalreport}
An~Yang, Baosong Yang, Beichen Zhang, Binyuan Hui, Bo~Zheng, Bowen Yu, Chengyuan Li, Dayiheng Liu, et~al.
\newblock Qwen2.5 technical report, 2025.
\newblock URL \url{https://arxiv.org/abs/2412.15115}.

\bibitem[Zhao et~al.(2018{\natexlab{a}})Zhao, Kim, Zhang, Rush, and LeCun]{zhao2018adversariallyregularizedautoencoders}
Jake Zhao, Yoon Kim, Kelly Zhang, Alexander~M. Rush, and Yann LeCun.
\newblock Adversarially regularized autoencoders, 2018{\natexlab{a}}.
\newblock URL \url{https://arxiv.org/abs/1706.04223}.

\bibitem[Zhao et~al.(2024)Zhao, Wang, Chen, Niu, Li, and U]{zhao_efficient_2024}
Kaiyan Zhao, Yiming Wang, Yuyang Chen, Xiaoguang Niu, Yan Li, and Leong~Hou U.
\newblock Efficient {Diversity}-based {Experience} {Replay} for {Deep} {Reinforcement} {Learning}, October 2024.
\newblock URL \url{http://arxiv.org/abs/2410.20487}.
\newblock arXiv:2410.20487 [cs] version: 1.

\bibitem[Zhao et~al.(2018{\natexlab{b}})Zhao, Song, and Ermon]{zhao2018infovaeinformationmaximizingvariational}
Shengjia Zhao, Jiaming Song, and Stefano Ermon.
\newblock Infovae: Information maximizing variational autoencoders, 2018{\natexlab{b}}.
\newblock URL \url{https://arxiv.org/abs/1706.02262}.

\bibitem[Zhao et~al.(2017)Zhao, Zhao, and Eskenazi]{zhao2017learningdiscourseleveldiversityneural}
Tiancheng Zhao, Ran Zhao, and Maxine Eskenazi.
\newblock Learning discourse-level diversity for neural dialog models using conditional variational autoencoders, 2017.
\newblock URL \url{https://arxiv.org/abs/1703.10960}.

\bibitem[Zheng et~al.(2024)Zheng, Tuyls, Peng, and Eysenbach]{zheng2024can}
Chongyi Zheng, Jens Tuyls, Joanne Peng, and Benjamin Eysenbach.
\newblock Can a misl fly? analysis and ingredients for mutual information skill learning.
\newblock \emph{arXiv preprint arXiv:2412.08021}, 2024.

\bibitem[Zhong et~al.(2025)Zhong, Shan, Feng, Xiong, Cheng, Zhao, He, Bian, and Wang]{zhong2025dpomeetspporeinforced}
Han Zhong, Zikang Shan, Guhao Feng, Wei Xiong, Xinle Cheng, Li~Zhao, Di~He, Jiang Bian, and Liwei Wang.
\newblock Dpo meets ppo: Reinforced token optimization for rlhf, 2025.
\newblock URL \url{https://arxiv.org/abs/2404.18922}.

\end{thebibliography}
\end{document}